\documentclass[final,5p,times,twocolumn]{elsarticle}

\usepackage{amsmath}
\usepackage{amssymb}
\usepackage{array}
\usepackage[caption=false,font=normalsize,labelfont=sf,textfont=sf]{subfig}
\usepackage{textcomp}
\usepackage{stfloats}
\usepackage{url}
\usepackage{verbatim}
\usepackage{graphicx}
\usepackage{cite}
\usepackage{booktabs}
\usepackage{tabularx}
\usepackage{multirow}
\begin{document}

\begin{frontmatter}



\title{
  Utilizing Autoregressive Networks for Full Lifecycle Data Generation of Rolling Bearings for RUL Prediction
}


\author[add1,add2]{Junliang Wang}

\author[add3]{Qinghua Zhang \corref{cor1}}
\ead{fengliangren@tom.com}
\cortext[cor1]{Corresponding author at: Guangdong Provincial Key Laboratory of Petrochemical Equipment and Fault Diagnosis, Guangdong
University of Petrochemical Technology, Maoming 525000, China.}
\author[add3]{Guanhua Zhu \corref{cor1}}
\ead{zhugh@gdupt.edu.cn}
\author[add3]{Guoxi Sun}

\affiliation[add1]{organization={School of Automation, Guangdong University
of Petrochemical Technology },%Department and Organization
            city={Maoming},
            postcode={525000}, 
            state={GuangZhou},
            country={China}}
\affiliation[add2]{organization={School
of Automation, Guangdong University of Technology },
            city={GuangZhou},
            postcode={510006}, 
            state={GuangZhou},
            country={China}}
\affiliation[add3]{organization={Guangdong Provincial Key Laboratory of Petrochemical Equipment and Fault
Diagnosis, Guangdong University of Petrochemical Technology },%Department and Organization
            city={Maoming},
            postcode={525000}, 
            state={GuangZhou},
            country={China}}
\begin{abstract}
The prediction of rolling bearing lifespan is of significant importance in industrial production. However, the scarcity of high-quality, full lifecycle data has been a major constraint in achieving precise predictions. To address this challenge, this paper introduces the CVGAN model, a novel framework capable of generating one-dimensional vibration signals in both horizontal and vertical directions, conditioned on historical vibration data and remaining useful life. In addition, we propose an autoregressive generation method that can iteratively utilize previously generated vibration information to guide the generation of current signals. The effectiveness of the CVGAN model is validated through experiments conducted on the PHM 2012 dataset. Our findings demonstrate that the CVGAN model, in terms of both MMD and FID metrics, outperforms many advanced methods in both autoregressive and non-autoregressive generation modes. Notably, training using the full lifecycle data generated by the CVGAN model significantly improves the performance of the predictive model. This result highlights the effectiveness of the data generated by CVGans in enhancing the predictive power of these models.
\end{abstract}


% \begin{highlights}
% \end{highlights}


\end{frontmatter}

\section{Introduction}
\label{sec:intro}
Rolling bearings are a crucial component of rotating machinery, and their performance tends to deteriorate over prolonged operation, potentially leading to machine failures. Consequently, predictive maintenance of these bearings is essential. A prevalent approach to bearing maintenance involves predicting the remaining useful life (RUL) of rolling bearings. In recent years, scholars from various countries have proposed numerous methods for predicting the RUL of rolling bearings. These methods can be categorized based on their approach: model-based 
\citep{physical1,physical2}, data-based, and hybrid methods \citep{hybrid1,hybrid2}. Model-based methods establish mathematical models of the bearing's life degradation process and analyze the failure mechanism mathematically to predict RUL. In contrast, data-based methods learn degradation patterns from sensor signals and establish predictive models using machine learning or deep learning. Hybrid methods aim to integrate the strengths of both physical models and data-driven approaches, thereby mitigating their limitations in RUL prediction.
In complex systems or under challenging operational conditions, obtaining physical models can be a significant hurdle. However, the rapid development of advanced sensing technologies has eased the acquisition of abundant industrial data, such as vibration signals. This wealth of data has paved the way for the application of deep learning \citep{7,8}. Consequently, various deep learning networks, such as recurrent neural networks, long short-term memory  networks, and their variants, have been widely utilized in the field of prognostics and health management (PHM)\citep{re5}.
For example, Li et al. \citep{1-1} proposed  a two-stage transfer regression CNN method, which first detects the incipient fault point and then predicts RUL.
Yang et al. \citep{1-2} introduced a method that incorporates uncertainty quantification, while Chen et al. \citep{1-3} and Guo et al. \citep{1-4} both proposed two-stage long short-term memory models.

In many instances, only a limited amount of data can be collected. Under such circumstances, the performance of network models significantly decreases. Therefore, it is imperative to explore life prediction methodologies that are effective in small sample scenarios.
Jin et al. \citep{small1} developed a reliability model for long-life satellite momentum wheels in small sample settings, using a physics-of-failure approach for lifespan prediction. A novel method using least squares support vector machine, optimized by particle swarm optimization, was proposed for predicting slewing bearing degradation with limited data \citep{small2}. Pan et al. \citep{small3} introduced a two-stage extreme learning machine method for efficient and accurate prediction of rolling-element bearings' RUL, particularly effective in small sample scenarios.
Chen et al. \citep{small4} introduced a novel approach using a deep convolutional neural network combined with a bidirectional long short-term memory network  and domain adaptation, significantly enhancing the accuracy and generalization of RUL predictions for equipment in small sample and complex condition scenarios.

In addition, the method of data generation is also an effective way to solve the problem of small samples.
Since the introduction of GAN\citep{gan} networks in 2014, numerous modifications and improvements have been made to enhance their performance and capabilities. One significant advancement was the wasserstein GAN (WGAN)\citep{wgan}, which employs the wasserstein distance as its loss function to increase training stability, enabling the model to learn in a more continuous and smoother manner. Following this development, Gulrajani et al. \citep{wgan-gp} introduced WGAN-GP (gradient penalty), a variant that incorporates a gradient penalty into the wasserstein distance, further stabilizing the training process.
Donahue et al. \citep{bigan} presented BiGAN, a model that, by integrating an encoder, not only generates data from latent space but also maps it back to this space. This architecture allows BiGAN to not only produce realistic data but also to learn effective representations of the data. Additionally, InfoGAN\citep{infogan} maximizes the mutual information between latent codes and observed data, facilitating the automatic learning of meaningful data representations. This approach enables InfoGAN to control and manipulate specific features in generated data, such as the angle or type of objects in images, demonstrating significant potential in image synthesis and feature interpretability.
Beyond image generation, GANs have also been utilized in the field of PHM for the generation of various vibration data.
Man et al.\citep{2-1} introduced a framework using AdCNN and CWGAN to generate high-quality training samples, improving prediction accuracy.
Wang et al.\citep{2-2}addressed the issue of unbalanced samples by combining wavelet transform denoising and Feature Preservation CycleGAN, enhancing feature extraction and prediction model construction.
Zhang et al.\citep{2-3} introduced a missing data generation method using an improved deep convolutional generative adversarial network, which enhances the stability and performance of RUL prediction models.

As an advancement of the autoencoder (AE) framework, the variational autoencoder (VAE)\citep{vae} has emerged as a significant tool in the realm of image generation. Characterized by its probabilistic approach, VAEs introduced a variational aspect to the encoding process, allowing for the generation of new images by sampling from the learned latent space distribution.
Peng et al.\citep{vae1}introduced a novel two-stage image inpainting model that leverages the hierarchical VQ-VAE\citep{vqvae} architecture, significantly improving the diversity and quality of inpainting results by separately generating and refining structural and textural information.
Li et al.\citep{vae2}proposed a bayesian generative model that integrates probabilistic generative models with neural networks, pretraining in a plain VAE manner.
Shi et al.\citep{vae3} introduced a multi-modal VAE that uses a mixture of experts  variational approach, characterizing successful learning in multimodal scenarios.
However, images generated by VAEs are often blurrier and less detailed compared to those produced by GANs, and VAEs are seldom used in the PHM domain. The CVAE-GAN network\citep{cvae-gan}, which combines the characteristics of conditional variational autoencoders (CVAE) and conditional generative adversarial networks (CGAN), aims to generate images of higher quality and diversity. This network structure leverages the stability of VAEs and the capacity of GANs to produce high-quality images. By introducing and improving the CVAE-GAN network, we have significantly enhanced the similarity and stability of directly generated vibration data for bearings.

Autoregressive methods, tools for modeling time series data, describe the relationship between a variable and its past observed values. These methods have particularly excelled in the field of Natural Language Processing (NLP). Chen et al. \citep{ar1} demonstrated state-of-the-art performance in both real-time and accuracy metrics. Autoregressive models are not only highly effective in NLP text generation tasks but have also recently gained traction in image generation. Brown et al.\citep{ar2} developed GPT-3, an autoregressive language model with 175 billion parameters, significantly larger than previous models, demonstrating exceptional performance in few-shot learning scenarios.

In the field of PHM, autoregressive methods are predominantly utilized for fault diagnosis tasks, often emphasizing physical models. However, Ma et al.\citep{ar2-2}proposed a novel hybrid model GNAR-GARCH that integrates linear and nonlinear autoregressive models with Generalized Autoregressive Conditional Heteroskedasticity. This model is especially adept at handling nonlinear and non-stationary signals, demonstrating enhanced accuracy and performance in the fault diagnosis of rolling bearings. Nistane et al.\citep{ar4} developed a health assessment model based on nonlinear autoregressive neural networks and health indicator extraction. This model employs friction torque sensors to capture vibration signals and utilizes continuous wavelet transform for feature extraction. It leverages NAR and NARX networks for predicting bearing degradation, showing high predictive accuracy.

\begin{figure*}
    \centering
    \includegraphics[width=1\textwidth]{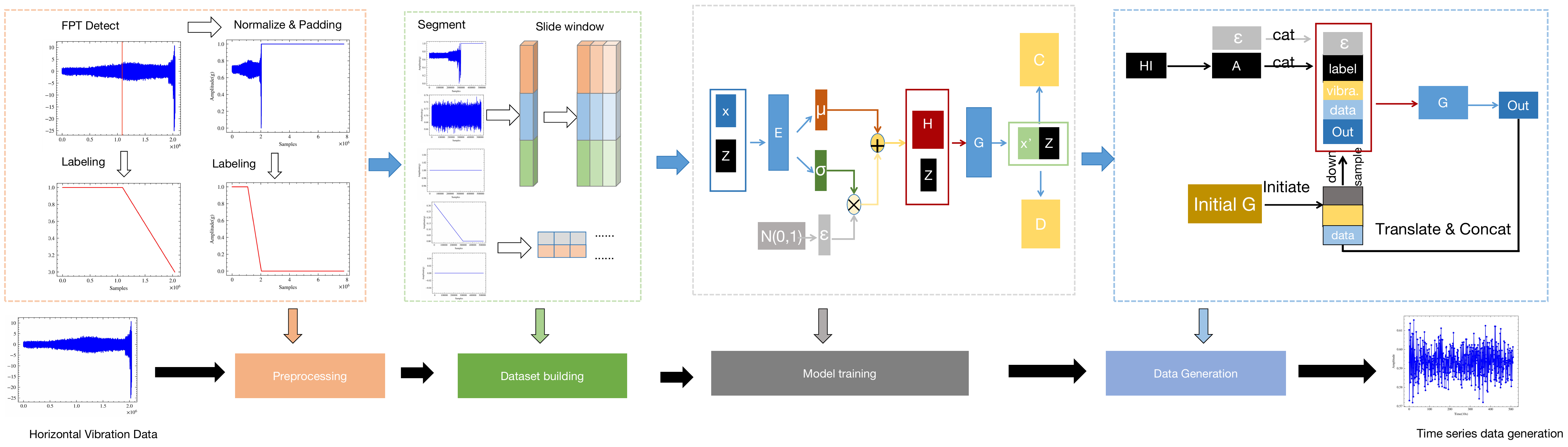}
    \caption{A flowchart of the proposed method.}
    \label{fig:overview}
\end{figure*}

Recognizing the nascent stage of autoregressive methods and full lifecycle data generation in the domain of lifespan prediction, we introduce a methodology capable of generating comprehensive lifecycle data through autoregression. This approach involves injecting categorical information to manipulate the duration of bearing life and the timing of fault onset. The operational workflow of our model is depicted in Figure \ref{fig:overview}.
The process begins with the normalization of all data, ensuring uniformity in numerical ranges. Subsequently, a dataset is constructed using segmentation and sliding window techniques. This dataset is then inputted into our proposed CVGAN model for training purposes. Concurrently, training is also required for an additional component, the initial generator. Upon completion of the training phase, both the initial generator and the main generator are employed for autoregressive generation.
The principal contributions of this paper are summarized into four key points:
\begin{enumerate}
   
 \item We introduce the CVGAN model, a novel framework based on CVAE and GAN. This model is capable of directly generating one-dimensional vibration signals in both horizontal and vertical directions, using historical vibration data and RUL as conditions.

 \item A unique autoregressive generation method is proposed, marking the first instance of generating full lifecycle vibration data for rolling bearings.

 \item The paper identifies and employs maximum mean discrepancy (MMD) and fréchet inception distance (FID) as evaluation metrics specifically for the task of generating one-dimensional vibration signals. This comparative analysis adds rigor to the assessment process.

 \item Experimental results on the PHM 2012 dataset demonstrate that both autoregressive and non-autoregressive methods of generation outperform many advanced techniques. Additionally, using the CVGAN model to generate complete lifecycle data significantly enhances the performance of multiple prediction models.
\end{enumerate}

The structure of the rest of the paper is as follows: Section \ref{sec:pre} provides a detailed introduction to GANs, the proposed CVGAN, and the autoregressive generation method. Section \ref{section:overview} describes the datasets used in the experiments, the creation of the training dataset, and some parameters used in training. Section \ref{section:Details} conducts a comprehensive evaluation of CVGAN and the autoregressive method through experimental analysis. Finally, Section \ref{section:con} concludes the paper and suggests potential directions for future research.

\section{Preliminary}
\label{sec:pre}
\subsection{GAN}
In a GAN architecture, there are two competing networks, the generator and the discriminator. The goal of the generator is to produce realistic data, while the task of the discriminator is to distinguish between the generated data and the real data. The two networks are pitted against each other during training, constantly improving their respective performance.

The goal of the discriminator is to correctly distinguish between real data and fake data generated by the generator. Its loss function is typically defined as a binary cross-entropy loss function, as the task of the discriminator can be viewed as a binary classification problem. For a given real sample and a generated sample, the discriminator loss function can be expressed as:

\begin{equation} \mathcal{L}_D =  -\mathbb{E}_{x \sim p_{\text{data}}(x)}[\log D(x)] - \mathbb{E}_{z \sim p_z(z)}[\log (1 - D(G(z)))] \end{equation}
Here, $P_{data}$ is the distribution of real data, $P_z
$ is the distribution of input noise for the generator, $D(x)$ is the output of the discriminator for a real sample $x$, and $G(z)$ is the output of the generator.

The objective of the generator is to fool the discriminator into mistaking its fake data for real data. The loss function of the generator can be formulated as:

\begin{equation} \mathcal{L}_G =  -\mathbb{E}_{z \sim p_z(z)}[\log D(G(z))] 
\end{equation}
Here, $P_z$ is the distribution of input noise for the generator, and $G(z)$ is the output of the generator.

These two loss functions together form the adversarial training framework of GANs. During training, the generator and discriminator are trained simultaneously. The generator tries to minimize $\mathcal{L}_G$, while the discriminator tries to minimize $\mathcal{L}_D$.This process is characterized by a continual improvement of capabilities on both sides, akin to a strategic game, where each party enhances its performance in response to the other's advancements. The ultimate goal of this iterative process is to reach a Nash equilibrium, a state where neither the generator nor the discriminator can further improve their strategies unilaterally, indicating a stable point in the training where the generator produces highly realistic outputs that the discriminator can barely distinguish from real data.
% $L_{\text{E}}= L_{\text{Recon}}+L_{\text{KL}}+L_{\text{Feature}}$
\subsection{CVGAN}
\begin{figure*}
    \centering
    \includegraphics[width=\textwidth]{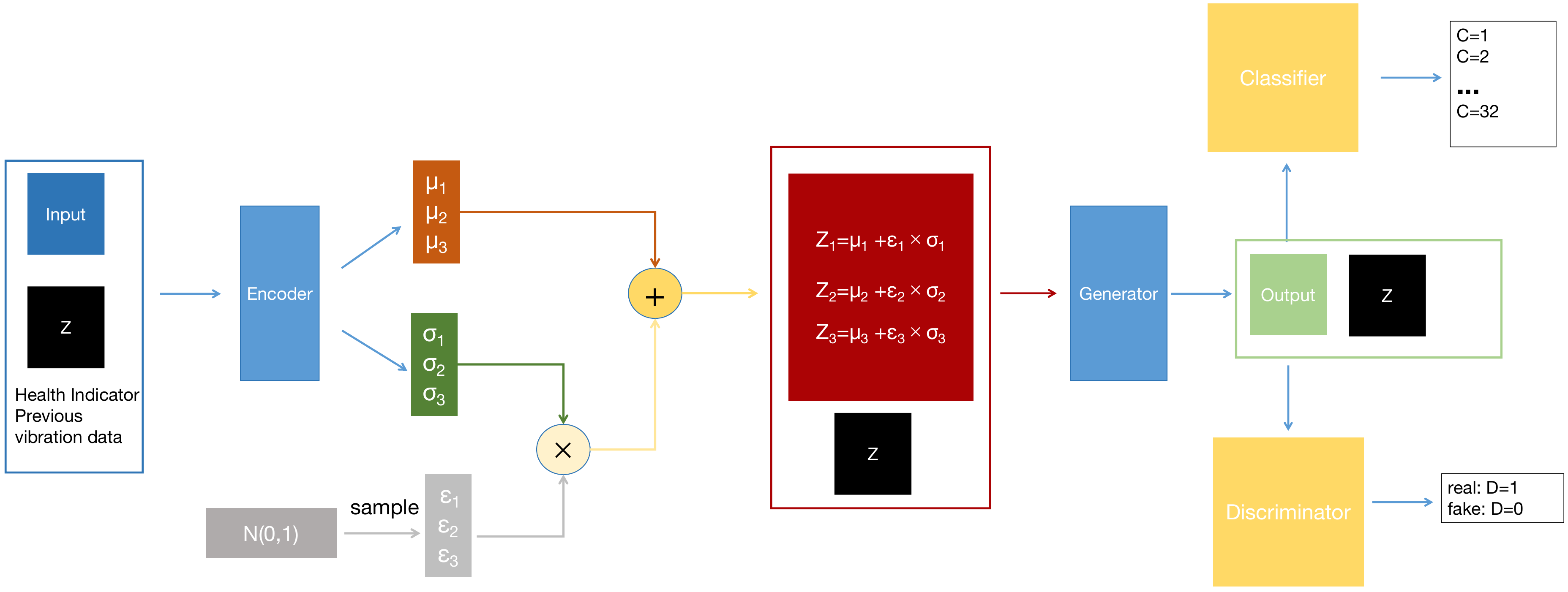}
    \caption{The basic structure of CVAE-GAN}
    \label{fig:cvgan}
\end{figure*}

In this paper, we introduce CVGAN as a generative model, which draws significantly from the structure of CVAE-GAN \citep{cvae-gan}. CVAE-GAN is an advanced network that integrates the features of CVAE and GAN. This model aims to harness the stability of CVAE and the generative capabilities of GAN to produce high-quality, conditionally constrained generated data.
The CVGAN consists of four main components: a CVAE, a generator, a discriminator, and a classifier, as shown in Figure \ref{fig:cvgan}. The VAE network within this structure uses an encoder to map input data to a latent space, which is then reconstructed into data by the decoder. As an enhancement of VAE, the CVAE can receive conditions as input. In our case, we use class health indicator and previous vibration data as conditions.

Meanwhile, the GAN component includes a generator, a discriminator, and a classifier. The generator is the aforementioned CVAE, while the discriminator attempts to distinguish between real and generated data, and the classifier aims to differentiate between various categories of generated images. These networks also receive conditioned inputs. By combining these frameworks, CVGAN ensures that the generator not only produces realistic data but also adheres to specific conditional constraints. In the context of CVAE-GAN, the decoder also functions as the generator in the GAN  component. Therefore, in subsequent paper, we do not differentiate between these two terms. Owing to the multiple components of CVGAN, its objective function can be combined in various ways. The objective function used in our paper is as follows:

\begin{align}
&\mathcal{L}_{\text{Recon}} = \| x - \hat{x} \|^2_2 \\
&\mathcal{L}_{\text{KL}} = D_{\text{KL}}(q(z|x,y) \| p(z)) \\
&\mathcal{L}_{\text{Feature}} = \|f_C(x) - f_C(\hat{x})\|^2_2 + \|f_D(x) - f_D(\hat{x})\|^2_2 \\
&\mathcal{L}_{\text{D}} = -\mathbb{E}_{x \sim p_{\text{data}}(x)}[\log D(x|y)] - \mathbb{E}_{z \sim p_z(z)}[\log (1 - D(G(z|y)))] \\
&\mathcal{L}_{\text{VAE}} = \mathcal{L}_{\text{Recon}} +\mathcal{L}_{\text{KL}}+ \mathcal{L}_{\text{Feature}}  \\
&\mathcal{L}_{\text{C}} = -\mathbb{E}_{x \sim p_{\text{data}}(x)}[\log C(x|y)]
\end{align}
Here, $x$ represents the real vibration data input, and $\hat x$ denotes the signal reconstructed by the VAE. $\mathcal{L}_{Recon}$ signifies the reconstruction loss, $\mathcal{L}_{KL}$ represents the Kullback-Leibler (KL) divergence loss, and $\mathcal{L}_{VAE}$ is the overall loss of the VAE. $\mathcal{L}_C$ denotes the classifier loss, while $\mathcal{L}_D$ stands for the discriminator loss. $D(x|y)$ indicates the discriminator's output, $C(x|y)$ signifies the classifier's output, $G(z|y)$ is the output of the generator, $f_C(x)$ refers to the intermediate features of the classifier, and $f_D(x)$ represents the intermediate features of the discriminator.
 
\subsubsection{Network Structure}
For the issue of bearing life prediction, we utilize the corresponding health indicator (HI) values as labels. Currently, there are two mainstream methods for constructing HI curves \citep{33,34}. One approach involves using an HI that linearly decreases from 1 to 0. The other method employs a piecewise function to simulate the degradation trend of bearings. In this case, the bearing's HI remains constant at 1 from the initial state until the first prediction time (FPT), after which it linearly decays.  Directly using the HI value as a condition for training might increase the learning burden on the generator. Therefore, we have divided the HI range from 0 to 1 into 32 categories. Concurrently, the objective of the classifier will also be to determine which of these 32 categories the generated vibration signal belongs to.

\begin{figure}
    \centering
    \includegraphics[width=\linewidth]{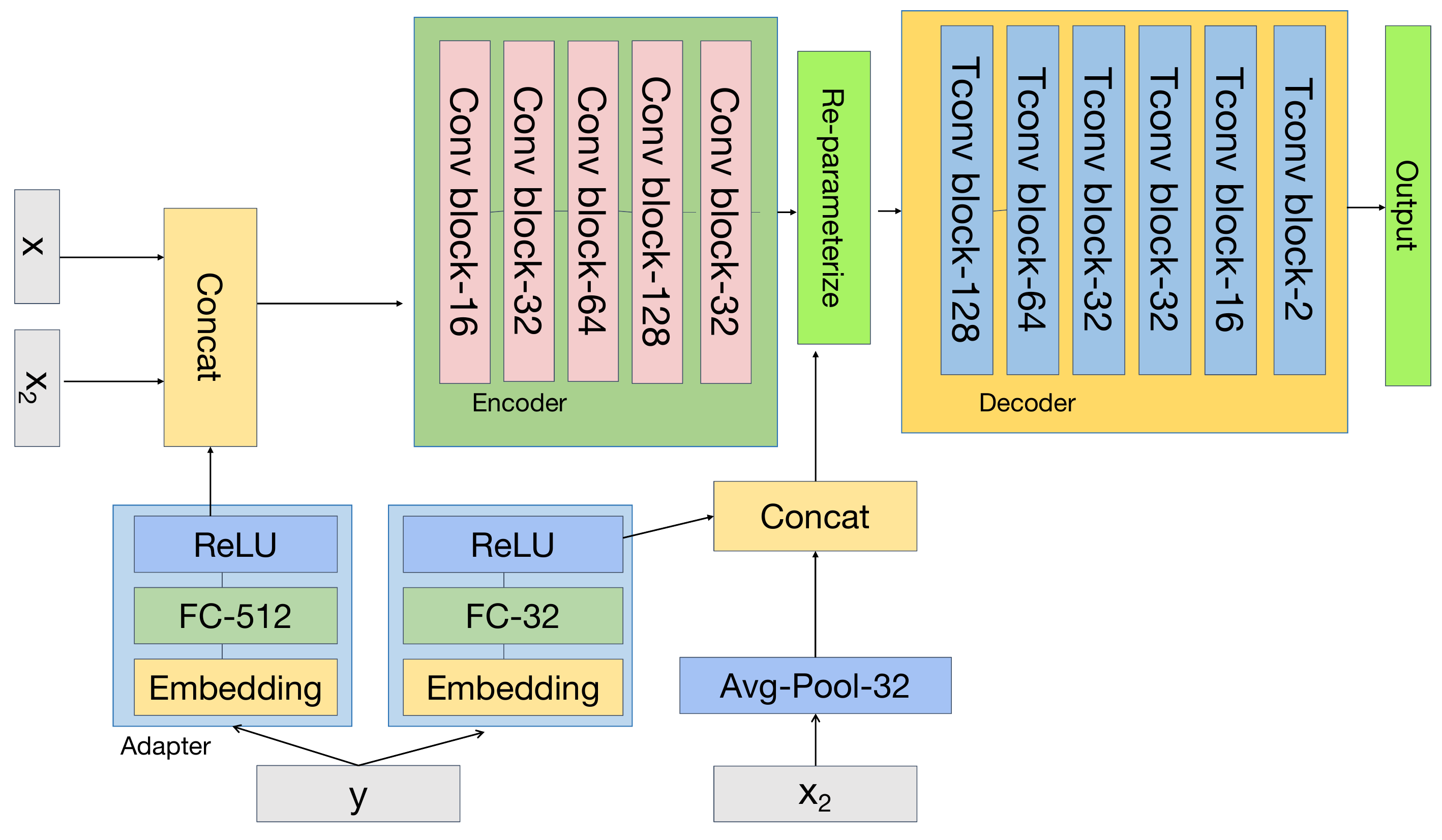}
    \caption{The specific network structure of VAE}
    \label{fig:vae}
\end{figure}

When inputting sample data, it is necessary to concatenate the original vibration data and conditions (including historical vibration data, category information) on the channel level. The specific workflow is illustrated in Figure \ref{fig:vae}. Upon input into the encoder, the category information $y$ first passes through an adapter to generate a 1x512 embedding. The historical vibration data $x_2$, with dimensions of $2k \times 512$ where k is typically 15, does not require adjustment when input into the encoder. After concatenation on the channel, the input dimension for the encoder is 512 x 33. After passing through the encoder and re-parameterization, a vector sampled from a normal distribution is generated, with a dimension of 1x32. When input into the decoder, the category information similarly passes through the Adapter to generate a 1x32 embedding. At this point, the historical vibration data also undergoes average pooling and downsampling to a size of 32. These conditional informations, along with the post-sampling vector, are then concatenated and fed into the decoder.

The network structure of the VAE is very simple. The specific architecture of the encoder is shown in Table \ref{tab:encoder}. The backbone network of the model consists of five convolutional blocks, each comprising a convolutional layer, a BN layer, and a LeakyReLU layer. In the first four convolutional blocks, the kernel size is 3x1, the stride is 2, and the padding is 1. After passing through a convolutional block, the size of the feature map is halved and the channels are doubled. The last convolutional block has a stride of 1 and channels are reduced to 32, followed by a sigmoid layer. 

The network structure of the generator is similar to that of the encoder, but uses transpose convolutional blocks and includes an additional block. The first four transpose convolutional blocks have a kernel size of 3x1, a stride of 2, and a padding of 1. After each transpose convolutional block, the size of the feature map doubles and the channels are halved. The last transpose convolutional block generates a two-channel vibration image, which is then passed to the discriminator and classifier.

The classifier and discriminator in our network also use a similar architecture to the encoder. However, the classifier's last convolutional block has 256 channels. This is followed by average pooling to reduce the feature map size to 1, and then sequentially through dropout and a fully connected  layer. The classifier's fully connected layer outputs information for 32 categories, while the discriminator's fully connected layer outputs a single classification indicating whether the image is real or not.

\begin{table} \caption{The parameters of the Encoder of VAE }  
\begin{tabularx}{\linewidth}{p{2cm}XXXp{2cm}} 
\toprule 
\textbf{Layer name} & \textbf{Kernels size} & \textbf{Stride} & \textbf{Kernel Num} & \textbf{Output size} \\ 
\midrule 
Input & – & – & – & 512 x 33 \\ 
Convolution1 & 1 x 3 & 1 x 2 & 16 & 256x 16 \\  Convolution2 & 1 x 3 & 1 x 2 & 32 & 128x 32 \\ 
Convolution3 & 1 x 3 & 1 x 2 & 64 & 64 x 64\\  
Convolution4 & 1 x 3 & 1 x 2 & 128 & 32 x 128 \\  Convolution5 & 1 x 3 & 1 x 1 & 32 & 32 x 32 \\ 
   \bottomrule \end{tabularx}  \label{tab:encoder} \end{table}

\subsection{Autoregressive Generation}
Autoregressive (AR) models are a tool used in time series data modeling to describe the relationship of a variable with its own past observations. In AR models, first-order autoregression (AR(1)) and second-order autoregression (AR(2)) are two common orders. The AR(1) model encapsulates the relationship between the current value and its immediate predecessor, while the AR(2) model extends this relationship to include the influence of the two preceding time points. Autoregressive models can be further extended to AR(p) models, where p denotes the order of the model, indicating the number of past consecutive observations included, including the current value. This allows for a more flexible capture of dependencies across different time points. The general mathematical form for an AR(p) model is: 
\begin{equation}
    X_t=c+\sum_{i=1}^p\phi_iX_{t-i}+\epsilon_t
\end{equation}
Here, $X_t$ represents the series value at time point $t$; $c$ is a constant term (which may be zero); $p$ denotes the order of the model, indicating the number of past consecutive observations included up to the current value; $\phi_i$ are the autoregressive coefficients, quantifying the impact of past observations on the current value; and $\epsilon_t$ is the error term, representing random disturbances.

\begin{figure}
    \centering
    \includegraphics[width=1\linewidth]{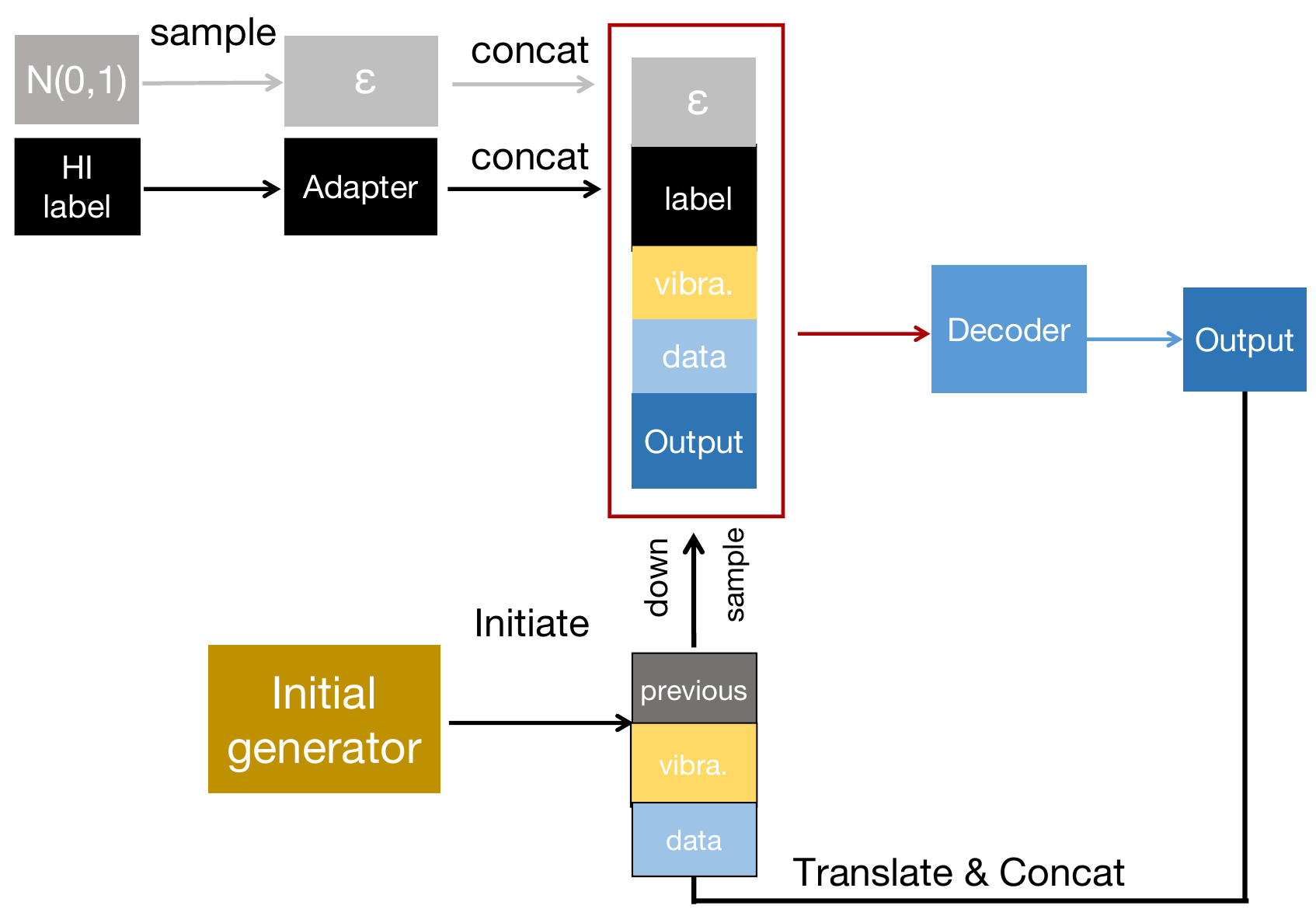}
    \caption{The proposed autoregressive generation method.}
    \label{fig:gen}
\end{figure}

Leveraging the advantageous characteristics of autoregressive models, we can effectively utilize historical vibration data to guide the generation of vibration data in the current window, thereby achieving continuous generation of time-series data. The specific generation principle, as shown in Figure \ref{fig:gen}, involves discarding the classifier, discriminator, and the encoder of the VAE during generation. Instead, we directly sample from a normal distribution and then restore it to a vibration signal through the decoder. The decoder receives category information and historical vibration data as conditions. The RUL category information is predetermined manually. Due to the initial absence of historical vibration data, we employ an initial generator to produce a window of historical vibration data. Subsequent historical vibration data is then generated by concatenating the current generated data with the existing historical data. Specifically, the size of the historical vibration data corresponds to the window size. In each cycle, the first data point is discarded, and the vibration signal generated in the current round is appended to the last, serving as input for the next round of generation.

\section{Experiments Overview} \label{section:overview}
\subsection{Dataset Introduction}  \label{section:dataset}
\begin{figure}
    \centering
    \includegraphics[width=1\linewidth]{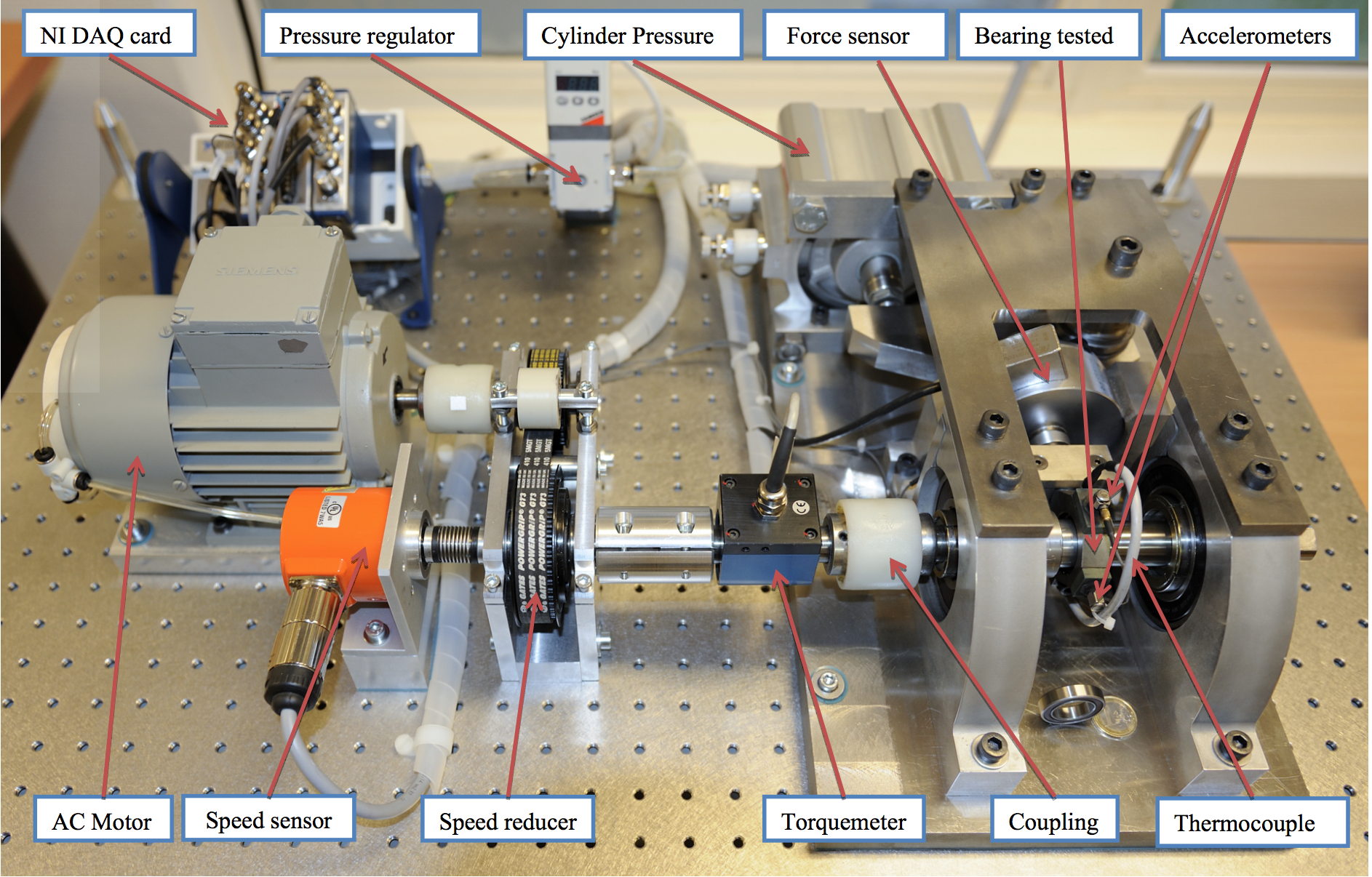}
    \caption{Overview of PRO-NOSTIA}
    \label{fig:dataset}
\end{figure}

As part of our research, we used a comprehensive dataset that tracks the performance of bearings from normal functionality through to failure, serving as a basis for testing our methodology \citep{dataset}. This particular dataset was first introduced during the IEEE PHM 2012 Data Challenge. The information contained within this dataset was gathered using the PRO-NOSTIA experimental setup, depicted in Figure \ref{fig:dataset}. Data acquisition was accomplished using two accelerometers, positioned horizontally and vertically. These devices were programmed to record at a sampling rate of 25.6 kHz, capturing data every 10 seconds, with each recording session lasting 0.1 seconds, thereby generating 2560 data points with each sample. To maintain test safety, if vibration readings surpassed 20 g, the experiment would be halted immediately. The moment when these vibration levels were exceeded was taken as the point of bearing failure. In this paper, we employ bearings under operating condition 1, characterized by a load capacity of 4000 N and a rotational speed of 1800 rpm. For determining the FPT point in the piecewise label, we adopted the method proposed by Li et al.\citep{fpt}, which is based on the 3$\sigma$ technique, as shown in Table \ref{tab:dataset}.

\begin{table} 
\caption{Full Life and FPT of Bearings under Condition 1 in the PHM 2012 Dataset} 
\centering 
\begin{tabularx}{\linewidth}{XXX} 
\toprule   
 \textbf{Bearing Number} & \textbf{FPT (s)} & \textbf{Actual Life (s)} \\
\midrule  
  Bearing1-1 & 11,420 & 28,030 \\ 
                         Bearing1-2 & 8,220  & 8,710 \\ 
                         Bearing1-3 & 9,600  & 23,750 \\ 
                         Bearing1-4 & 10,180 & 14,280 \\ 
                        Bearing1-5 & 24,070 & 24,630 \\ 
                         Bearing1-6 & 16,270 & 24,480 \\ 
                         Bearing1-7 & 22,040 & 22,590 \\  
\bottomrule 
\end{tabularx}  
\label{tab:dataset} 
\end{table}

\subsection{Dataset Build}
In this paper, we organize the vibration signal data of bearings into a series: $v_1,v_2,…,v_{n-1},v_n$, where each value is paired with a corresponding HI, specifically including $z_1,z_2,…,z_{n-1},z_n$. The HI, as utilized here, comprises the piecewise label that were detailed in Section \ref{section:dataset}.
For the raw data, after normalization, we apply average pooling to each sample to ensure uniform feature dimensions across different datasets and operating conditions of bearings. This process transforms the dimension of each $v_i$ to $2N_{feature}$, where $N_{feature}$ represents the number of features.
We then construct historical data using a sliding window approach with a window size of $k$. This means that every set of $k$ lengths of vibration data is used as a sample. Since the vibration data has two channels—horizontal and vertical—the dimension of the historical data $x_2$ is $k \times 2N_{feature}$.
For the first window, the data from $v_1$ to $v_k$ serves as the historical data $x_2$, and the corresponding label is $z_{k+1}$. Another input to the model, $x$, is $v_{k+1}$ with a dimension of $2N_{feature}$. The window is then shifted one time unit along the sampling time axis to repeat this process and generate a series of input samples. From each segment, $n-k$ samples can be obtained. The last window is not used for data construction due to the lack of a corresponding label.

\subsection{Parameter Setting}
During the training process of the generative model, the AdamW optimizer is employed to minimize the loss function. The learning rate for the encoder and generator is set at 0.0006, while for the discriminator and classifier, it is set at 0.0002. The batch size is 1024. The network is trained for 100 epochs, with an early stop setting of 30 epochs. In the training of the predictive model, the same AdamW optimizer is used with a learning rate of 0.0008 and a batch size of 2048. This network is trained for 150 epochs, with an early stop set at 20 epochs. The results presented in this paper are the averages of five independent runs using random seeds 15, 25, 35, 45, and 55, to ensure consistency and replicability. All experiments are conducted on an NVIDIA GeForce 3090 (24GB) GPU and implemented using the PyTorch framework.

For the generative model, all bearings from condition 1 are used as the training set. The predictive model employs an alternative method for dividing the test set; specifically, one of the seven available bearings is selected as the test set, while the remaining six bearings serve as the training set. In this paper, bearings 1-1 and 1-3 are primarily used as test sets. Consequently, bearings 1-2 to 1-7 and bearings 1-1, 1-2, 1-4 to 1-7 are used as training sets, respectively.

\section{Experiments Details} \label{section:Details}
\subsection{Selection of Metrics for Generation Method}
In the task of generating 1-dimensional vibration signal time series, the academic community has not yet established a unified evaluation metric. Therefore, we have selected a series of indicators commonly used in the literature on bearing vibration signal generation, including MMD (maximum mean discrepancy), FID (Fréchet inception distance), PSNR (peak signal-to-noise ratio), MTD (mean tarvel distance), and MV (mean variance)\citep{our2}, as illustrated in Table \ref{tab:metrics}. For these metrics, through experimental validation, we have identified the most accurate and reliable indicators for subsequent analysis. 

In the evaluation of generative models, MMD is often employed to measure the discrepancy between the distribution of generated data and the true data distribution. A lower MMD value indicates closer statistical characteristics between the generated and real data. In the context of vibration data, to reduce the computational time of MMD, we divide it into horizontal and numerical channels. Each channel's data is dimensionally reduced to 64 dimensions using principal component analysis  and then mapped using a gaussian kernel function. 

FID is a popular metric for assessing the quality of generated images, especially in image generative models like GANs. It operates by comparing the distribution of generated images to real images in a deep feature space. Typically, this involves extracting features using the Inception network and calculating the Fréchet distance between these two distributions. A lower FID score usually indicates higher image quality. However, since the Inception model is not applicable for vibration signal inspection, we trained a classification model as a substitute. 

Given that normal-phase bearing vibration data maintains a stable mean value, we also utilized this characteristic to calculate the mean absolute difference (MAD) between the mean values of the generated data set and the original data set during the normal phase. This is the most intuitive method for comparing differences between the source bearing and the generated counterpart, although it becomes less accurate with smaller values. We also evaluated the generated quality using the peak signal-to-noise ratio (PSNR) metric. PSNR quantifies the quality of reconstruction by comparing the discrepancies between the original and reconstructed images. Additionally, we employed the mean travel distance (MTD) and mean variance (MV) methods to assess the quality of predicted vibration data. MTD computes the absolute distance between each pair of adjacent points, summing these distances and averaging the result. MV calculates the variance of data within each window and subsequently averages the variances across all windows. The mean squared error (MSE) metric was used to compare the predictive results with the category labels used as conditions, where the differences are squared and then averaged.

\begin{table*}[htbp]
\centering % Center the table
\caption{Different evaluation metric equations} % Caption for the table
\label{tab:metrics}
\renewcommand{\arraystretch}{2}
\begin{tabularx}{\textwidth}{XXX}
\toprule
\textbf{Index} & \multicolumn{2}{c}{\textbf{Formula}}  \\
\midrule
id1& $\begin{array}{l} \bar x_i  = \frac{1}{M}\sum_{j=1}^{M} x_{i, j} \end{array}$&  $\begin{array}{l}\text{MAD}  = \frac{1}{N}\sum_{i=1}^{N}|\bar {\hat x_i}-\bar x_{i}| \end{array}$ \\ 
id2 &
\multicolumn{2}{l}{
    $\begin{array}{l}
        K(x,y)=\exp(-\| x-y\|^2) \\
        \text{MMD} = \mathbb{E}_{x,x^{\prime}\sim p_{g}}[K(x,x^{\prime})] - 2\mathbb{E}_{x\sim p_{g},\hat x\sim p_{data}}[K(x,\hat x)] + \mathbb{E}_{\hat x,\hat x^{\prime}\sim p_{data}}[K(\hat x,\hat x^{\prime})]
    \end{array} $
} \\
id3 &
\multicolumn{2}{l}{
$
\text{FID}  = \|\mu_{f(x)} - \mu_{f(\hat x)}\|^2 + \text{Tr}(\Sigma_{f(x)} + \Sigma_{f(\hat x)} - 2(\Sigma_{f(x)} \Sigma_{f(\hat x)})^{1/2})$} 
 \\ 

 id4 &
 \multicolumn{2}{l}{
$
\text{MSE}  = \frac{1}{N}\sum_{i=1}^{N}(y_i-P(x_{i}))^2 $ } \\

id5 &
\multicolumn{2}{l}{
 $
 \text{MTD}  = \frac{1}{n-1}\sum_{i=2}^n|P(x_i)-P(x_{i-1})| $} \\

id6 &
    $\begin{array}{l}
        e = \frac{1}{MN} \sum_{i=1}^N \sum_{j=1}^M (x_{i, j} - \hat x_{i, j})^2\end{array}$   &
        $\begin{array}{l}
        \text{PSNR} = 10 \cdot \log_{10} \left( \frac{\text{MAX}_I^2}{e} \right)
    \end{array}$  \\
 id7 &

    $\begin{array}{l}
        \bar x_i = \frac{1}{k} \sum_{j=1}^{k} P(x_{i+j}) \\
        \sigma_i = \frac{1}{k}\sum_{j=0}^{k-1}(P(x_{i+j})-\bar x_i)^2 \end{array}$ &
        $\begin{array}{l}
        \text{MV} = \frac{1}{N-k+1}\sum_{i=1}^{N-k+1}\sigma_i 
        \end{array}$
        \\
\bottomrule
\end{tabularx}%
\flushleft
\footnotesize % Make the font smaller for the note
Here, \( x_i \) denotes the generated signal of the \( i^{\text{th}} \) sample, while \( \hat{x}_i \) represents the original \( i^{\text{th}} \) vibration signal. The term \( y_i \) refers to the RUL value corresponding to the \( i^{\text{th}} \) vibration signal. \( P(x) \) is the output of the predictive model. The function \( f(x) \) yields the intermediate features produced by the classification model. The symbol \( \mu_{f(x)} \) denotes the mean value of these features across the second dimension, and \( \Sigma_{f(x)} \) represents the covariance of these features. \( \text{MAX}_I \) is the maximum value among all vibration signals, and \( N \) represents the total number of samples.
\end{table*}

\begin{table}[htbp]
\centering
\caption{The comparison results of different statistical indicators}
\label{tab:conf}
% \resizebox{\columnwidth}{!}{%
\begin{tabular}{@{}llll@{}}
\toprule
\textbf{Metrics}   & \textbf{Conf1} & \textbf{Conf2}              & \textbf{Conf3}             \\ \midrule
Generator Loss  $\downarrow$   & 194.32          & 10.28                        & \textbf{3.54}                        \\
Discriminator Loss $\downarrow$ & 185.54          &  15.51                     &  \textbf{9.01} \\
Horizontal MMD $\downarrow$     & 1.4050          & 0.4819                       & \textbf{0.0007}                      \\
Vertical MMD $\downarrow$      & 0.7419          & 0.7526                       & \textbf{0.0025}                      \\
FID $\downarrow$               & 143.30          & 22.79                        & \textbf{4.23}                        \\
Horizontal MAD $\downarrow$    & 0.0871          & 0.1437                       & \textbf{0.0017}                      \\
Vertical MAD $\downarrow$    & 0.0484          & 0.0297                       & \textbf{0.0014}                      \\
MSE $\downarrow$               & 108.89          & 121.66                       & \textbf{64.10}                       \\
MV  $\downarrow$               & 22.56           & 16.73                        & \textbf{6.29}                        \\
MTD $\downarrow$               & 6.02            & 4.64                         & \textbf{3.18}                        \\
PSNR $\downarrow$              & 20.92           & \textbf{18.21}                        & 33.68                       \\ \bottomrule
\end{tabular}%
% }
\end{table}

\begin{figure*}[htbp]
    \centering
    \begin{tabular}{cc}
        \includegraphics[height=4.5cm]{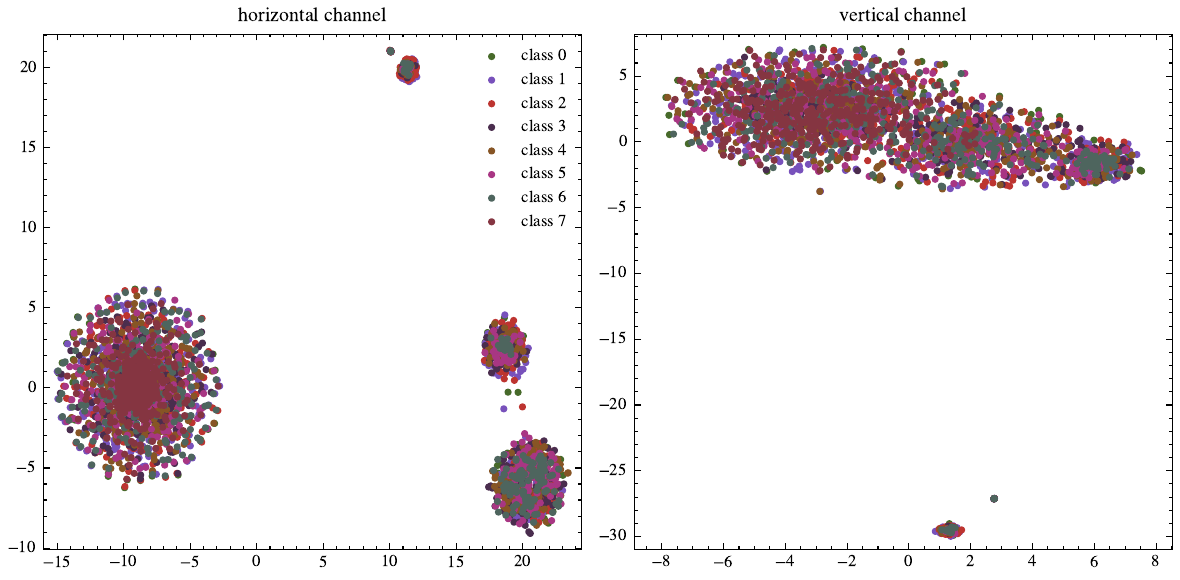} & \includegraphics[height=4.5cm]{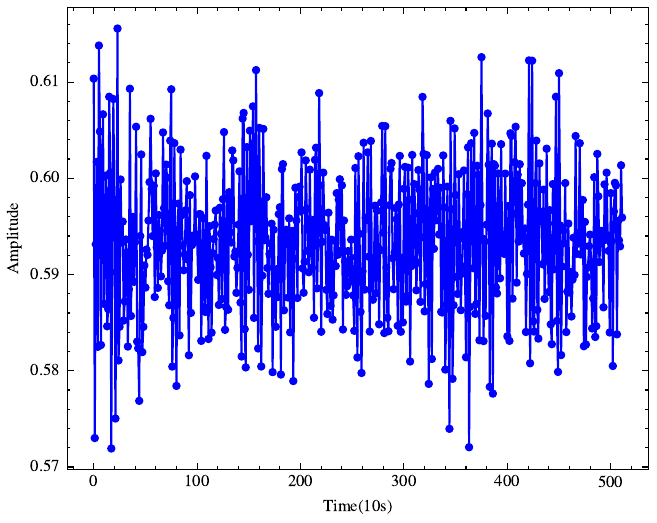} \\
        (a) t-SNE & (b) vibration data \\
    \end{tabular}
    \caption{Integrated Analysis of real data  }
    \label{fig:contrast}
\end{figure*}
\begin{figure*}[htbp]
    \centering
    \begin{tabular}{cc}
        \includegraphics[height=4.5cm]{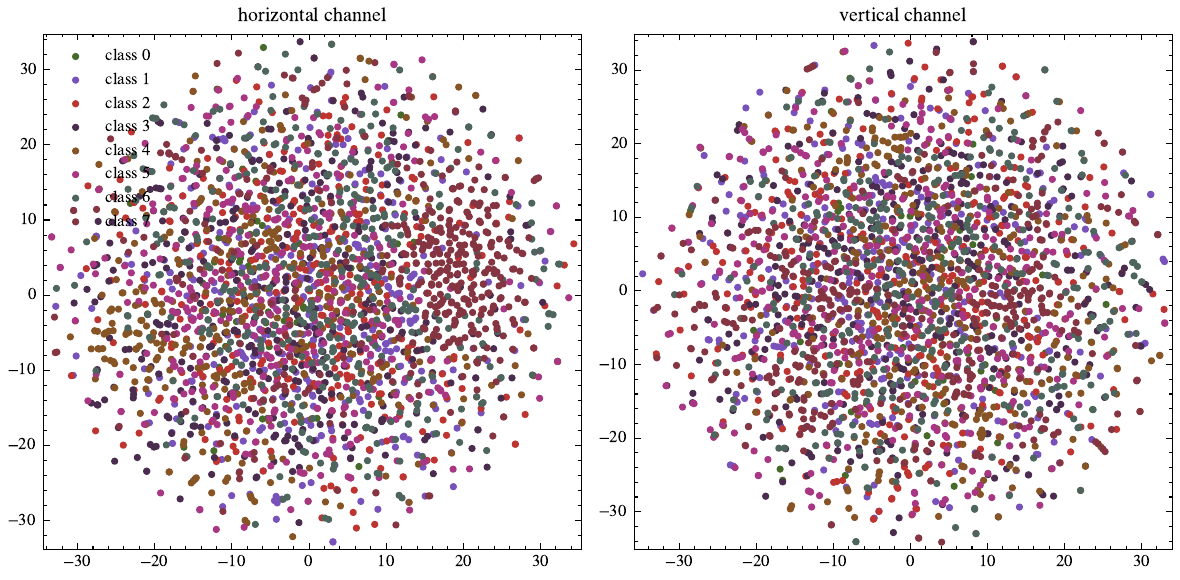} & \includegraphics[height=4.5cm]{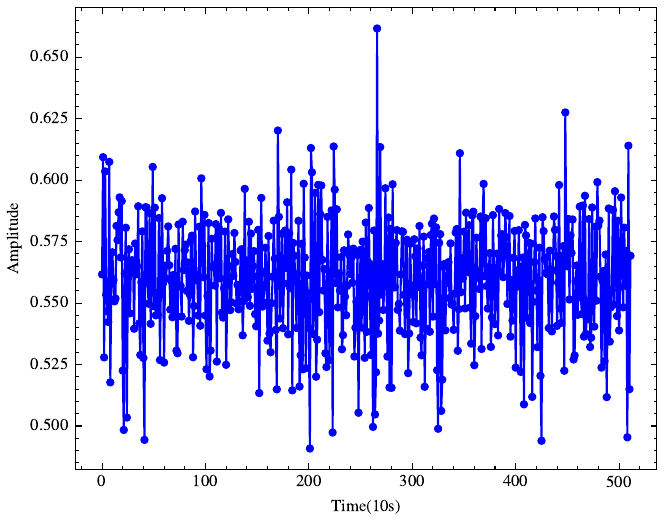} \\
        (a) t-SNE & (b) vibration data \\
    \end{tabular}
    \caption{Integrated Analysis of Hyperparameter Configurations 1 on Generation Data  }
    \label{fig:conf1}
\end{figure*}
\begin{figure*}[htbp]
    \centering
    \begin{tabular}{cc}
        \includegraphics[height=4.5cm]{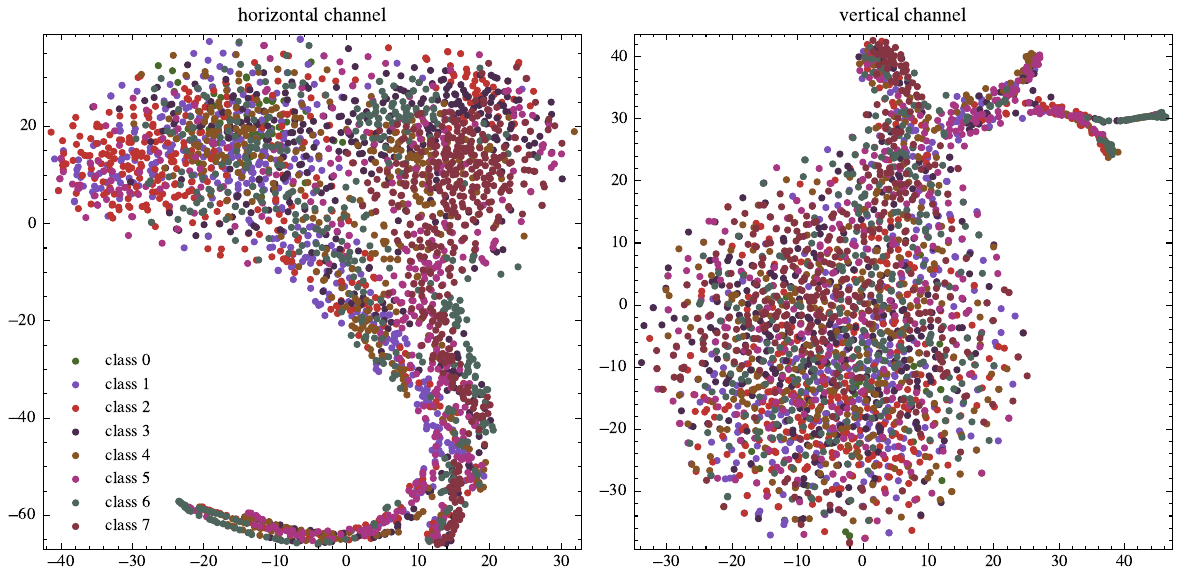} & \includegraphics[height=4.5cm]{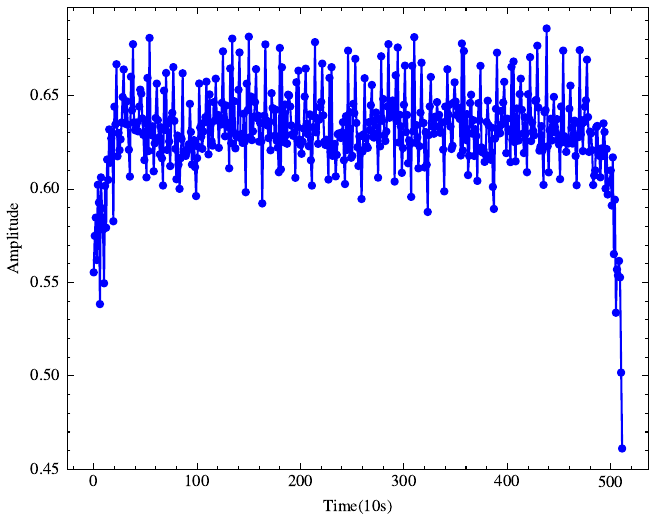} \\
        (a) t-SNE & (b) vibration data \\
    \end{tabular}
    \caption{Integrated Analysis of Hyperparameter Configurations 2 on Generation Data }
    \label{fig:conf2}
\end{figure*}
\begin{figure*}[htbp]
    \centering
    \begin{tabular}{cc}
        \includegraphics[height=4.5cm]{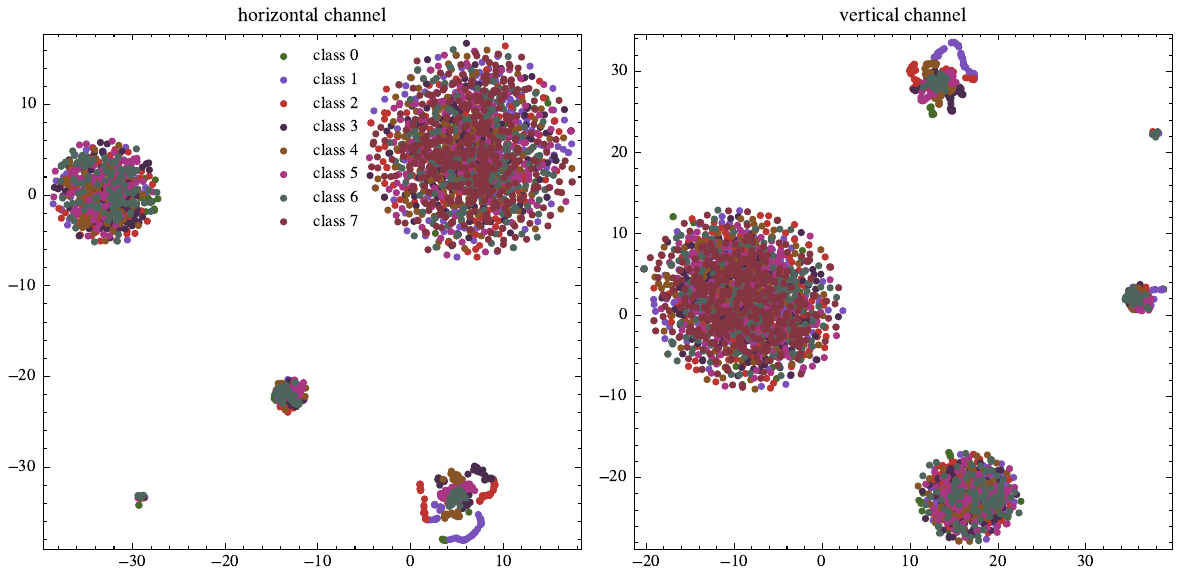} & \includegraphics[height=4.5cm]{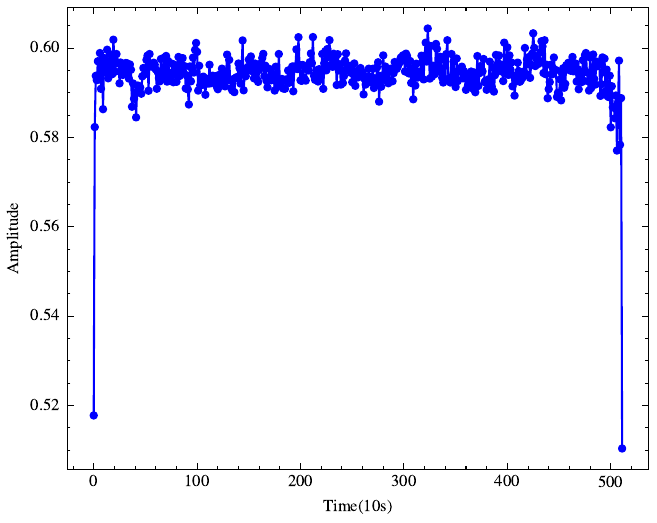} \\
        (a) t-SNE & (b) vibration data \\
    \end{tabular}
    \caption{Integrated Analysis of Hyperparameter Configurations 3 on Generation Data  }
    \label{fig:conf3}
\end{figure*}
In analyzing these metrics, we employed t-Distributed stochastic neighbor embedding (t-SNE) and training loss as preliminary tools to assess the applicability of the aforementioned indicators. In order to display clearly and beautifully on the t-SNE picture, we divide the original 32 categories of HI into 8 categories, and select 400 samples for each category.
Initially, we selected three different hyperparameter configurations for our model. The generative effects and t-SNE results of these configurations are depicted in the figures. For Figures \ref{fig:conf1} and \ref{fig:conf2}, the scatter degree in the t-SNE of both channels has decreased, indicating an improvement in generation quality. However, the vibration signal generated in subfigure c might not accurately reflect this improvement.
Regarding Figure \ref{fig:conf3}, the t-SNE analysis shows a more concentrated distribution of generated data, with the horizontal t-SNE result closely resembling the four cluster centers of the original distribution. The mean value of the vibration signal is around 0.59, aligning more closely with the mean of the original data.

Table \ref{tab:conf} illustrates the losses for configurations 1 to 3, clearly indicating a gradual increase in generative quality from Figures 1 to 3. This information is useful for analyzing different statistical metrics. However, we observed that some of these metrics, such as MSE, PSNR, and horizontal MAD , surprisingly show an increasing trend. The MV and MTD metrics, despite significant optimization from configurations 1 to configurations 3, do not exhibit substantial variability in values. horizontal MAD and vertical MAD, which directly compare the mean value differences during the normal phase, are relatively rudimentary comparison methods, hence we decided not to use them for model evaluation.

Through this straightforward quantitative approach, we have identified two viable evaluation metrics: MMD and FID. Both of these metrics measure differences between distributions and are indeed effective in reflecting the discrepancies between generated vibration signals and actual signals.

\subsection{Network Structure Selection}
In this section, we explore the impact of various training methodologies and loss functions on the outcomes of autoregressive generation. We first examined whether autoregressive training, is beneficial for autoregressive generation. Our approach involved selecting training methods that align with autoregressive generation, such as matched autoregressive training, standard non-autoregressive training followed by autoregressive fine-tuning (limited to a single epoch), and conventional non-autoregressive training. The experimental results of these methods are presented in Table \ref{tab:train_method}. Notably, non-autoregressive training demonstrated a significant lead over other models, an unexpected finding that underscores the relative underdevelopment of current autoregressive training techniques compared to the optimization results of larger batch sizes in standard methods. Furthermore, autoregressive training yielded the poorest results among all the methods, inferior even to the combination of standard training followed by autoregressive fine-tuning. The three fine-tuning methods showed similar performance in terms of MMD and FID, with a slight edge in MMD for fine-tuning only the VAE. This observation reinforces our earlier conclusion about the inadequacy of autoregressive training methods. Therefore,  we employ non-autoregressive training with autoregressive generation as the main framework of our model.

To investigate the impact of different loss functions on autoregressive results, we defined several configurations with a focus on improving the generative quality of VAEs, as shown in Table \ref{tab:loss_item} and Table \ref{tab:loss_config}.  In addition to the loss functions delineated in Section 2.2, our study introduces several specialized loss functions. The $\mathcal{L}_{mf}$ loss function is designed to align the generated images with the original data based on the average intermediate features generated by the discriminator. The $\mathcal{L}_{mc}$ function aligns the generated images with the original data based on the class-center features produced by a classifier, enhancing categorical accuracy.

Further, the $\mathcal{L}_{he}$ loss function aligns the average historical vibration data with the reconstructed results. The $\mathcal{L}_{hp}$ function similarly aligns the average historical vibration data but with the directly generated images, focusing on maintaining consistency between historical data and new generative outputs.

Lastly, $\mathcal{L}_{Bin}$ is inspired by the traditional generator loss function in GANs. Instead of relying on intermediate features, this function directly utilizes the outputs from the classifier and discriminator, maintaining a traditional approach in loss calculation.

\begin{figure*}
    \centering
    \includegraphics[width=1\textwidth]{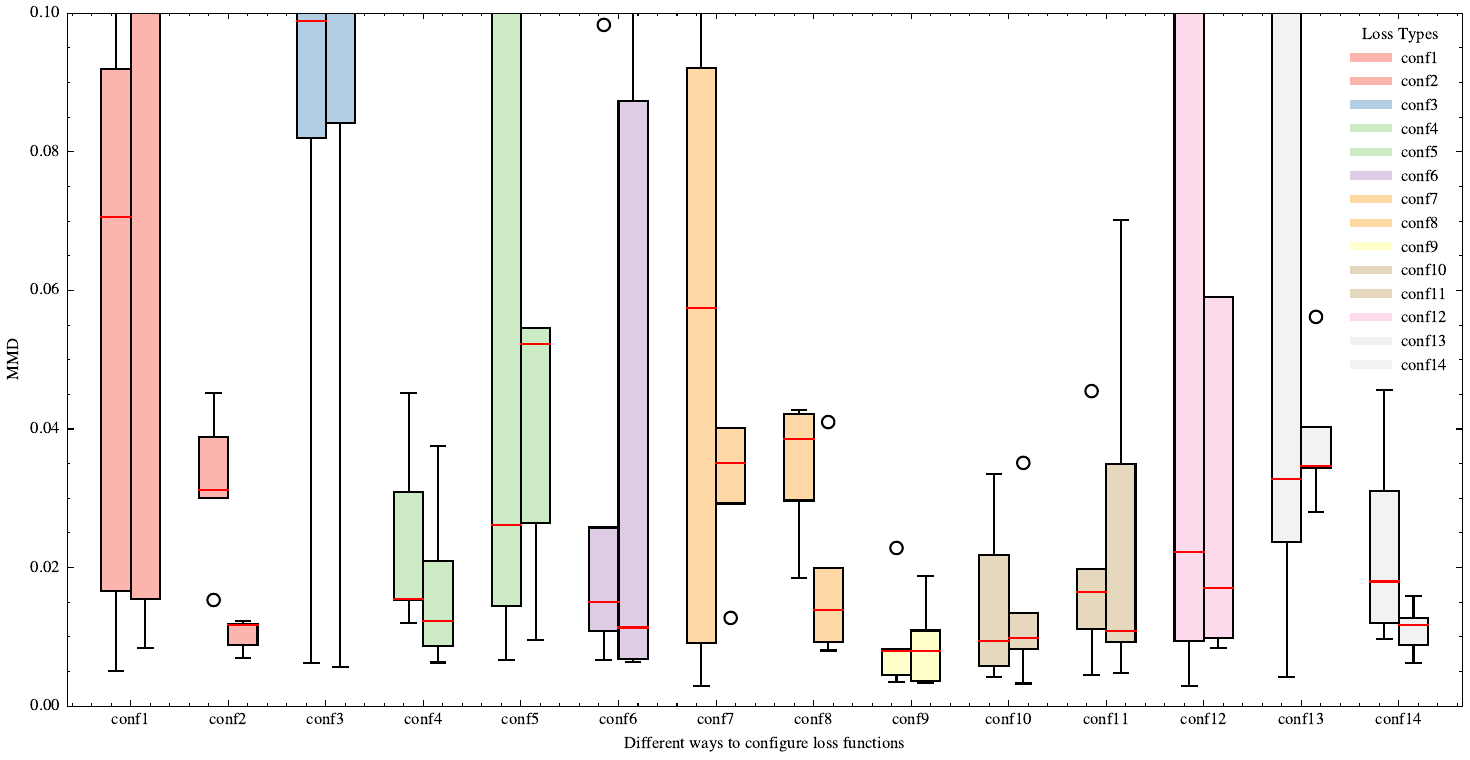}
    \caption{Evaluation results of different loss function configurations}
    \label{fig:loss}
\end{figure*}

Models trained with these different loss functions underwent autoregressive inference to generate images, and their MMD test results are presented in Figure \ref{fig:loss}. Notably, the configuration labeled 'conf9' demonstrated optimal performance in both horizontal and vertical aspects of MMD. Without imposing additional constraints and using simple configurations, it achieved the best results. However, when we added various other loss terms to 'conf9', such as in 'conf14' with the inclusion of $\mathcal{L}_{mc}$ (Loss of middle classifier), intending for the middle features of the classifier to closely align with each category for real and generated data, the horizontal MMD increased. When compared with the classic loss function of GANs, specifically with 'conf1', the results showed significant improvements in generation quality with reduced fluctuations. The classic loss only achieved smaller fluctuations in configurations 'conf10' and 'conf11', which included $\mathcal{L}_{mc}$ and $\mathcal{L}_{mf}$. This suggests that in the context of autoregressive generation, traditional GAN loss provides insufficient guidance for the continuity of generated time-series images, and a more explicit approach, such as aligning features, is required. The experimental results demonstrate that using $\mathcal{L}_{feature}$ in place of $\mathcal{L}_{Bin}$ significantly enhances both the quality and stability of the generated results. Curiously, a comparison between 'conf1' and 'conf3', 'conf7' and 'conf10' reveals that the addition of the $\mathcal{L}_1$ constraint to the discriminator led to a significant regression in generation quality. Theoretically, it is indeed necessary to train the discriminator with reconstructed vibration data. This paradoxical outcome poses a question for future research.

\begin{table}[htbp]
\centering
\caption{Evaluation results of different training methods}
\label{tab:train_method}
\resizebox{\columnwidth}{!}{%
\begin{tabular}{@{}llll@{}}
\toprule
\textbf{Method}                                        & \textbf{Horizontal MMD}$\downarrow$ & \textbf{Vertical MMD}$\downarrow$ & \textbf{FID} $\downarrow$         \\ \midrule
Non-AR Training                      & \textbf{0.006}    & \textbf{0.008}  & \textbf{3.304} \\
AR Fine-tuning w/o Discriminator \& Classifier & 0.112    & 0.039  & 7.082 \\
AR Fine-tuning w/o Classifier                 & 0.116    & 0.042  & 6.971 \\
Comprehensive AR Fine-tuning                  & 0.117    & 0.042  & 6.862 \\
AR Training                                   & 0.181    & 0.033  & 13.508 \\ \bottomrule
\end{tabular}%
}
\end{table}

\begin{table*}[b]
\centering
\caption{
Different loss function configuration item equations}
\label{tab:loss_item}
\resizebox{\textwidth}{!}{
\begin{tabular}{llll}
\toprule
\textbf{Index} & \textbf{Formula} & \textbf{Index} & \textbf{Formula} \\ \midrule
id1 & $
 \mathcal{L}_{\text{Recon}} = \| x - \hat{x} \|^2_2
 $ & id7 &
 $
\mathcal{L}_{\text{KL}} = D_{\text{KL}}(q(z|x) \| p(z))
 $ \\
id2 & $\mathcal{L}_{he}=\frac{1}{2}||\frac{1}{k}\sum_{i}^{k}x_2-\hat{x}||_{2}^{2}$ & id8 & 
 $
\mathcal{L}_{\text{Feature}} = \|f_C(x) - f_C(\hat{x})\|^2_2+ \|f_D(x) - f_D(\hat{x})\|^2_2
 $ \\ 
 id3 &
$\mathcal{L}_{mf}=\frac{1}{2}||\frac{1}{m}\sum_{i}^{m}f_{D}(x)-\frac{1}{m}\sum_{i}^{m}f_{D}(G(z|y))||_{2}^{2}$ & id9 & $\mathcal{L}_{hp}=\frac{1}{2}||\frac{1}{k}\sum_{i}^{k}x_2-G(z|y)||_{2}^{2} $ \\
id4 & $\mathcal{L}_{1}= - \mathbb{E}[\log (1 - D(\hat{x}|y))]$ & id10 & $\mathcal{L}_{Bin}= -\mathbb{E}[\log ( D(\hat{x}|y))]-\mathbb{E}[\log(\text{C}(\hat{x}|y))]$ \\

id5& $\mathcal{L}_{\text{C}} = -\mathbb{E}_{x \sim p_{\text{data}}(x)}[\log(\text{C}(x|y))]$ & id11 & $\mathcal{L}_{mc}=\frac{1}{2}\sum_{c_{i}}||f_{C}^{c_{i}}(x)-f_{C}^{c_{i}}(G(z|y))||_{2}^{2}$\\
id6 & \multicolumn{3}{l}{
 $\mathcal{L}_{\text{d}} = -\mathbb{E}_{x \sim p_{\text{data}}(x)}[\log D(x|y)] - \mathbb{E}_{z \sim p_z(z)}[\log (1 - D(G(z|y)))]$}  \\
 \bottomrule
 \end{tabular}%
 }
 \flushleft
\footnotesize % Make the font smaller for the note
Here, $x$ represents the real vibration data input,  $ \hat x$ denotes the signal reconstructed by the VAE, $x_2$ denotes the real historical vibration data. The term \( y \) refers to the conditions of the input model. $D(x|y)$ indicates the discriminator’s output, $C(x|y)$ signifies the classifier’s output, $E(x|y)$ is the output of the encoder and $G(z|y)$ is the output of the generator. The function \( f_C(x) \) yields the intermediate features produced by the classification model. The function \( f_D(x) \) yields the intermediate features produced by the discriminator model. $f_{C}^{c_{i}}$ represents the exponential moving average result for each class i of the intermediate features produced by the classification model. $k$ denotes the window size and $m$  represents the dimension size of intermediate features produced by the discriminator model.
\end{table*}

\begin{table}[htbp]
\centering % Center the table
\caption{Different loss function configuration.} 
\label{tab:loss_config}
\renewcommand{\arraystretch}{1.5}
\resizebox{\linewidth}{!}{
\begin{tabular}{@{}clcl@{}}
\toprule
\textbf{Index} & \textbf{Loss Items} &\textbf{Index} & \textbf{Loss Items} \\
\midrule
 Conf1& $\begin{array}{ll}
  \mathcal{L}_{VAE}= \mathcal{L}_{\text{Bin}}\\
  \mathcal{L}_{D}= \mathcal{L}_d
   \end{array}$ & Conf8 &  
$\begin{array}{ll}\mathcal{L}_{VAE}= \mathcal{L}_{\text{Feature}}+\mathcal{L}_{\text{mc}}+\mathcal{L}_{\text{mf}}\\
 \mathcal{L}_{D}= \mathcal{L}_d  +\mathcal{L}_1
 \end{array}$\\
   Conf2 &
 $\begin{array}{ll}\mathcal{L}_{VAE}= \mathcal{L}_{\text{Feature}}+\mathcal{L}_{\text{mc}}+\mathcal{L}_{\text{hp}}\\
 \mathcal{L}_{D}= \mathcal{L}_d 
 \end{array}$     &  Conf9 &  
$\begin{array}{ll}\mathcal{L}_{VAE}= \mathcal{L}_{\text{Feature}}\\
 \mathcal{L}_{D}= \mathcal{L}_d  
 \end{array}$ \\
  Conf3 & $\begin{array}{ll}
  L_{VAE}= \mathcal{L}_{\text{Bin}}\\
  \mathcal{L}_{D}= \mathcal{L}_d +\mathcal{L}_1
  \end{array}$    &  Conf10 &  
$\begin{array}{ll}\mathcal{L}_{VAE}= \mathcal{L}_{\text{Bin}}+\mathcal{L}_{\text{mc}}+\mathcal{L}_{\text{mf}}\\
 \mathcal{L}_{D}= \mathcal{L}_d  
 \end{array}$ \\

Conf4 &
 $\begin{array}{ll}\mathcal{L}_{VAE}= \mathcal{L}_{\text{Feature}}+\mathcal{L}_{\text{mc}}+\mathcal{L}_{\text{hp}}+\mathcal{L}_{\text{he}}\\
 \mathcal{L}_{D}= \mathcal{L}_d 
 \end{array}$& Conf11 &  
$\begin{array}{ll}\mathcal{L}_{VAE}= \mathcal{L}_{\text{Bin}}+\mathcal{L}_{\text{mc}}+\mathcal{L}_{\text{mf}}+L_{\text{hp}}\\
 \mathcal{L}_{D}= \mathcal{L}_d  
 \end{array}$\\
 Conf5 &
   $\begin{array}{ll}\mathcal{L}_{VAE}=\mathcal{L}_{\text{bin}}+ \mathcal{L}_{\text{mc}}\\
 \mathcal{L}_{D}= \mathcal{L}_d +\mathcal{L}_1
 \end{array}$ &Conf12 &  
$\begin{array}{ll}\mathcal{L}_{VAE}= \mathcal{L}_{\text{Bin}}+\mathcal{L}_{\text{mc}}\\
 \mathcal{L}_{D}= \mathcal{L}_d  
 \end{array}$\\ 
Conf6 &  
$\begin{array}{ll}\mathcal{L}_{VAE}= \mathcal{L}_{\text{bin}}+\mathcal{L}_{\text{mc}}+\mathcal{L}_{\text{mf}}+\mathcal{L}_{\text{he}}\\
 \mathcal{L}_{D}= \mathcal{L}_d 
 \end{array}$  & Conf13 &  
$\begin{array}{ll}\mathcal{L}_{VAE}= \mathcal{L}_{\text{Bin}}+\mathcal{L}_{\text{mc}}+\mathcal{L}_{\text{hp}}\\
 \mathcal{L}_{D}= \mathcal{L}_d  
 \end{array}$\\

 Conf7 & $ 
\begin{array}{ll}\mathcal{L}_{VAE}= \mathcal{L}_{\text{bin}}+\mathcal{L}_{\text{mc}}+\mathcal{L}_{\text{mf}}\\
 \mathcal{L}_{D}= \mathcal{L}_d  +\mathcal{L}_1
 \end{array}$ &
 Conf14 &  
$\begin{array}{ll}\mathcal{L}_{VAE}= \mathcal{L}_{\text{Feature}}+\mathcal{L}_{\text{mc}}\\
 \mathcal{L}_{D}= \mathcal{L}_d  
 \end{array}$\\

 \bottomrule
\end{tabular}%
}
\flushleft
\footnotesize
The $\mathcal{L}_{VAE}$ inherently includes the two terms $\mathcal{L}_{\text{Recon}}$ and $\mathcal{L}_{\text{KL}}$. Therefore, we do not list them separately in the table.
\end{table}

\begin{table}[]
\centering
\caption{The results of ablation experiment.}
\label{tab:ablation}
\resizebox{\columnwidth}{!}{%
\begin{tabular}{@{}llll@{}}
\toprule
\textbf{Models} & \textbf{Horizontal MMD} $\downarrow$ & \textbf{Vertical MMD} $\downarrow$ & \textbf{FID} $\downarrow$ \\ \midrule
CGAN            & 0.5746                  & 0.5730                & 13.671       \\
CVAE            & 0.0080                  & 0.0104                & 4.275        \\
CVGAN           & \textbf{0.0044}                  & \textbf{0.0080}                & \textbf{2.778}        \\
CVGAN\_no\_H    & 0.9209                  & 1.3527                & 94.150       \\
GAN             & 0.6312                  & 0.5865                & 116.184      \\
VAE             & 1.8299                  & 1.7844                & 98.472       \\
VGAN            & 1.4294                  & 1.4164                & 117.530      \\ \midrule 
\addlinespace[5pt]
                 \multicolumn{4}{c}{\textbf{Non-Autoregressive}}                         \\ \addlinespace[5pt]
                
CGAN            & 0.2050                  & 0.3061                & 26.446       \\
CVAE            & 0.0009                  & 0.0027                & \textbf{3.664}        \\
CVGAN           & \textbf{0.0007}                  & \textbf{0.0025}                & 4.225        \\
CVGAN\_no\_H    & 0.9224                  & 1.3517                & 94.760       \\
GAN             & 0.6318                  & 0.5837                & 118.229      \\
VAE             & 1.8302                  & 1.7820                & 99.161       \\
VGAN            & 1.4315                  & 1.4154                & 118.967      \\ \bottomrule
\end{tabular}%
}
\end{table}

\subsection{Ablation Study}
In this section, we conducted ablation experiments on our model, testing and evaluating each component individually. The model will be split into the following structures: 
\begin{enumerate}
\item CGAN: Utilized the same discriminator and classifier as our model, but the generative module was replaced with the decoder from traditional GANs. The generator, discriminator, and classifier in CGAN were all capable of receiving conditional inputs. The purpose of this model is to explore the effect of the generator on the result.
\item GAN and CGAN: Maintained a similar structure, but with reduced channel numbers. All networks within these models did not accept conditional inputs. 
\item CVAE: Employed the generative part of our original model, lacking a discriminator and classifier, but capable of receiving conditional inputs. The purpose of this model is to explore the effects of discriminators and classifiers on the results.
\item VAE and CVAE: Had identical structures, but VAE had reduced channel numbers and its networks did not accept conditional inputs. 
\item VGAN: Shared a similar structure with our method but had reduced channel numbers, with none of the networks accepting conditional inputs. 
\item CVGAN no-H: Mirrored our method's structure but only accepted categorical conditions, not historical vibration data as input. 
\end{enumerate}

\begin{figure*}[htbp]
    \centering
    \begin{tabular}{cc}
        \includegraphics[width=0.45\textwidth]{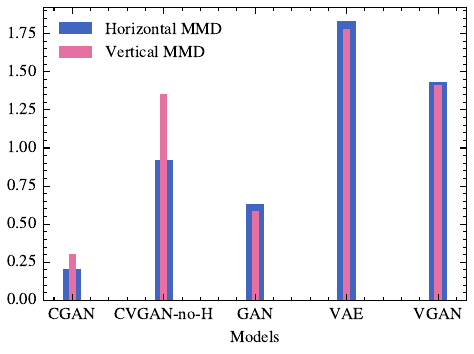} & \includegraphics[width=0.45\textwidth]{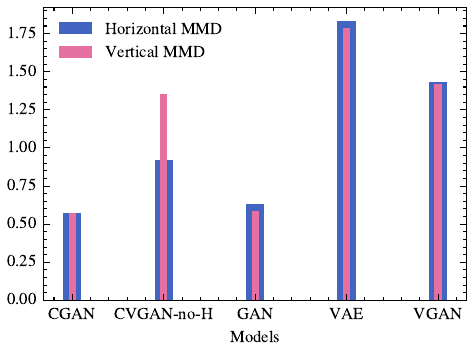} \\
        (a) NAR Generation & (b) AR Generation \\
    \end{tabular}
    \caption{MMD results of ablation experiment  }
    \label{fig:ablation}
\end{figure*}

From Figure \ref{fig:ablation}and Table \ref{tab:ablation}, it's evident that removing historical vibration data as input significantly increased the MMD for CVGAN no-H compared to CVGAN, underscoring the importance of historical vibration information in enhancing generative quality. Without historical vibration data, the difficulty of generation increased. CVGAN no-H, GAN, VAE, and VGAN, lacking historical vibration data input, showed negligible differences between autoregressive and non-autoregressive generation. VGAN, without conditional input, performed worse than a standard GAN. The original VAE achieved the poorest results, but CVAE, with added conditional information, showed significant improvement. This highlights the importance of incorporating labels and historical vibration data as conditions, similar to the difference between GAN and CGAN. CVAE performed second-best after our model, significantly outperforming CGAN, suggesting that an encoder-decoder generative network is superior to a purely decoder-based structure. While CVGAN showed limited improvement over CVAE in terms of MMD and FID metrics, its actual generative quality was substantially better.

\subsection{Comparsion with other Models}
In this section, we benchmarked our model against some of the most advanced models in the field, including InfoGAN\citep{infogan}, BiGAN\citep{bigan}, WGAN\citep{wgan}, WGAN-GP\citep{wgan-gp}, CVAE, and AAE-GAN\citep{aae-gan}, to compare their autoregressive and non-autoregressive generative capabilities.
\begin{enumerate}
\item InfoGAN: We primarily adhered to its original setup. InfoGAN is a conditional GAN, but since its discriminator is required to output latent code information, we omitted the historical vibration data as a conditional input.

\item BiGAN: We used its original setup. BiGAN is an unconditional GAN, which means it doesn't take additional conditional inputs for its generative process.

\item WGAN and WGAN-GP: These models largely inherited the network structure of CGAN. However, they stayed close to the original papers in terms of loss functions, optimizers, and learning rates. WGANs are known for their stable training dynamics due to the Wasserstein loss and gradient penalty (in WGAN-GP).

\item AAE-GAN: Also set up according to its original configuration, AAE-GAN uses an AE as the architecture for its generator. Its discriminator doesn't judge the final generated images but rather the latent variables. Since a randomly sampled normal distribution serves as the true label in our implementation, the discriminator is unconditional, whereas the generator accepts category and historical vibration data as conditions.
\end{enumerate}

\begin{figure*}[htbp]
    \centering
    \begin{tabular}{cc}
        \includegraphics[width=0.45\textwidth]{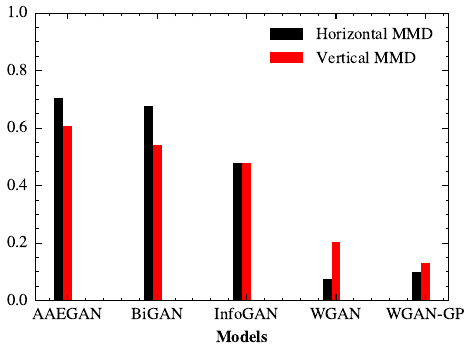} & \includegraphics[width=0.45\textwidth]{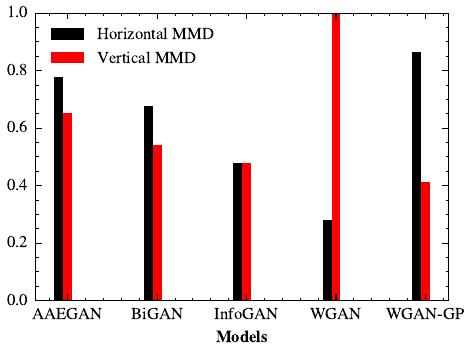} \\
        (a) NAR Generation & (b) AR Generation \\
    \end{tabular}
    \caption{MMD result for different models}
    \label{fig:compare}
\end{figure*}

Figure \ref{fig:compare}a illustrates the effectiveness of non-autoregressive generation, highlighting that the performance of the CVAE significantly surpasses that of the other five models. The detailed results of the CVAE, as shown in Table \ref{tab:compare}, indicate its superior performance. For clarity, these results are not depicted in Figure \ref{fig:compare}. It is observed that the generative outcomes of WGAN and WGAN-GP are markedly superior compared to the other three models. However, the MMD along the horizontal axis remains at 0.2, indicating that the quality of generation still has room for improvement. Notably, both the InfoGAN and the BiGAN are constrained by their inability to incorporate historical oscillation data as input, which increases the complexity of generation and results in inferior performance relative to models that accept such conditions. Surprisingly, the AAEGAN demonstrates the least effective performance, unexpectedly falling short of even unconditional generative models. This could potentially be attributed to inherent design issues within the AAEGAN model itself.

Figure \ref{fig:compare}b depicts the outcomes of generative models utilizing autoregressive methods. A notable observation is that after applying autoregressive generation, the MMD values for both WGAN and WGAN-GP show a significant increase, indicating the presence of some errors in autoregressive generation. The models InfoGAN and BiGAN, which are incapable of utilizing historical oscillation data as input, exhibit almost no difference in their generative results between autoregressive and non-autoregressive methods. This finding aligns with the conclusions drawn from the ablation experiments in the previous section. In contrast, the AAEGAN, due to its generator's ability to process historical oscillation data, demonstrates a slight increase in MMD.
Based on these observations, it can be concluded that a discriminator equipped with historical oscillation and category conditions can significantly enhance the quality of generation. However, it is also more susceptible to the influences of autoregressive processes.

\begin{table}[htbp]
    \centering
    \caption{The results of different models}
    \label{tab:compare}
    \resizebox{\columnwidth}{!}{
    \begin{tabular}{lccc}
    \toprule
        \textbf{Models} & \textbf{Horizontal MMD}$\downarrow$ & \textbf{Vertical MMD}$\downarrow$ & \textbf{FID}$\downarrow$ \\ \midrule
        AAEGAN & 0.7781  & 0.6525  & 141.048  \\ 
        BIGAN & 0.6779  & 0.5427  & 146.951  \\ 
        CVAE & \textbf{0.0080}  & \textbf{0.0104}  & \textbf{4.275}  \\ 
        INFOGAN & 0.4794  & 0.4803  & 117.916  \\ 
        WGAN & 0.2795  & 1.1391  & 10.961  \\ 
        WGAN-GP & 0.8639  & 0.4142  & 17.433  \\ \midrule
       \addlinespace[5pt]
                 \multicolumn{4}{c}{\textbf{Non-Autoregressive}}                         \\ \addlinespace[5pt]
        AAEGAN & 0.7062  & 0.6094  & 75.607  \\ 
        BIGAN & 0.6787  & 0.5415  & 149.029  \\ 
        CVAE & \textbf{0.0009}  & \textbf{0.0027}  & \textbf{3.664}  \\ 
        INFOGAN & 0.4779  & 0.4790  & 118.335  \\ 
        WGAN & 0.0744  & 0.2033  & 14.041  \\ 
        WGAN-GP & 0.1012  & 0.1323  & 29.179 \\ \bottomrule
    \end{tabular}}
\end{table}

\subsection{Autoregressive Evalation}
In this section, we evaluated the full effect of autoregressive generation. Specifically, we employed an unconditional model to generate a set of vibration data for a specified window, which was subsequently used as historical data for the autoregressive model. The experiment was configured with the total lifespan of the generated bearings set at 1000, and a FPT threshold at 300. We selected the best-performing models identified in previous discussions, namely the CVGAN, CVAE, and WGAN, for the final generation tests.

\begin{figure*}[htbp]
    \centering
    \begin{tabular}{ccc}
        \includegraphics[height=3.8cm]{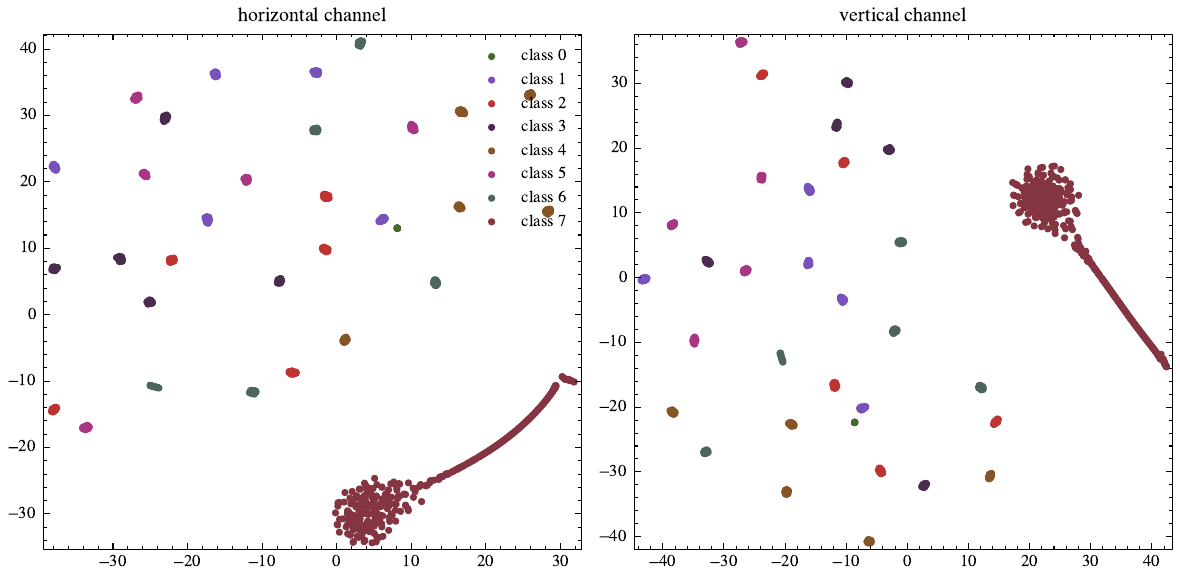} & 
        \includegraphics[height=3.8cm]{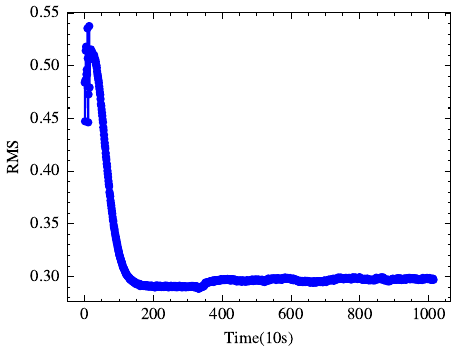} & \includegraphics[height=3.8cm]{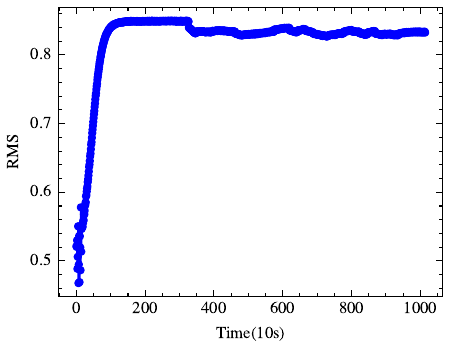} \\
        (a) t-SNE & (b) Horizontal RMS (c) Vertical RMS\\
    \end{tabular}
    \caption{Autoregressive Generation Evaluation Results of WGAN}
    \label{fig:pred1}
\end{figure*}

\begin{figure*}[htbp]
    \centering
    \begin{tabular}{ccc}
        \includegraphics[height=3.8cm]{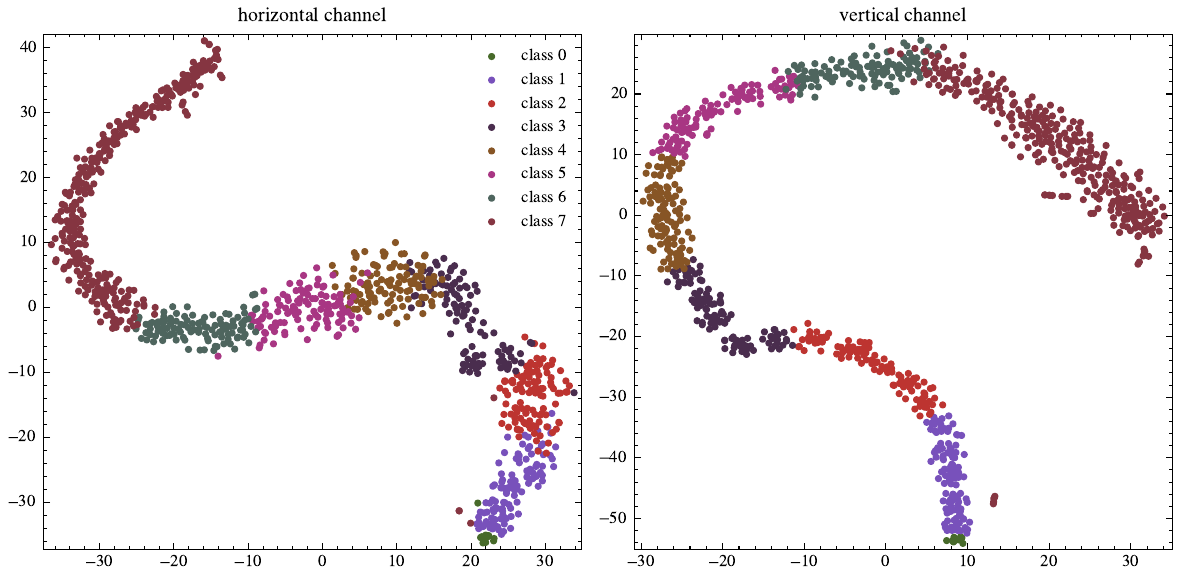} & \includegraphics[height=3.8cm]{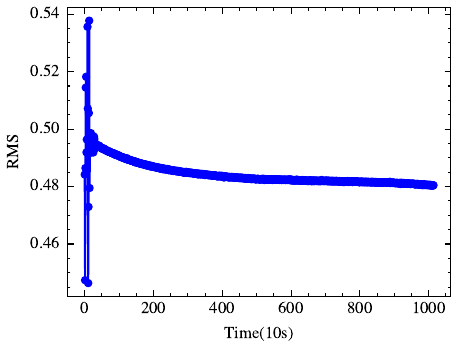} & \includegraphics[height=3.8cm]{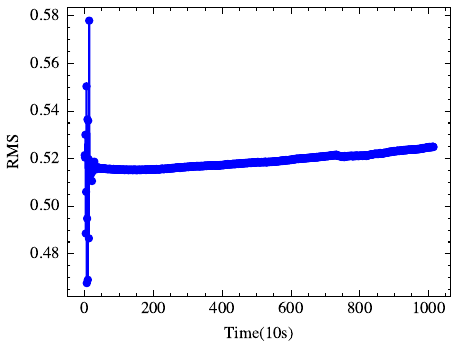} \\
        (a) t-SNE & (b) Horizontal RMS (c) Vertical RMS\\
    \end{tabular}
    \caption{Autoregressive Generation Evaluation Results of CVAE}
    \label{fig:pred2}
\end{figure*}

\begin{figure*}[htbp]
    \centering
    \begin{tabular}{ccc}
        \includegraphics[height=3.8cm]{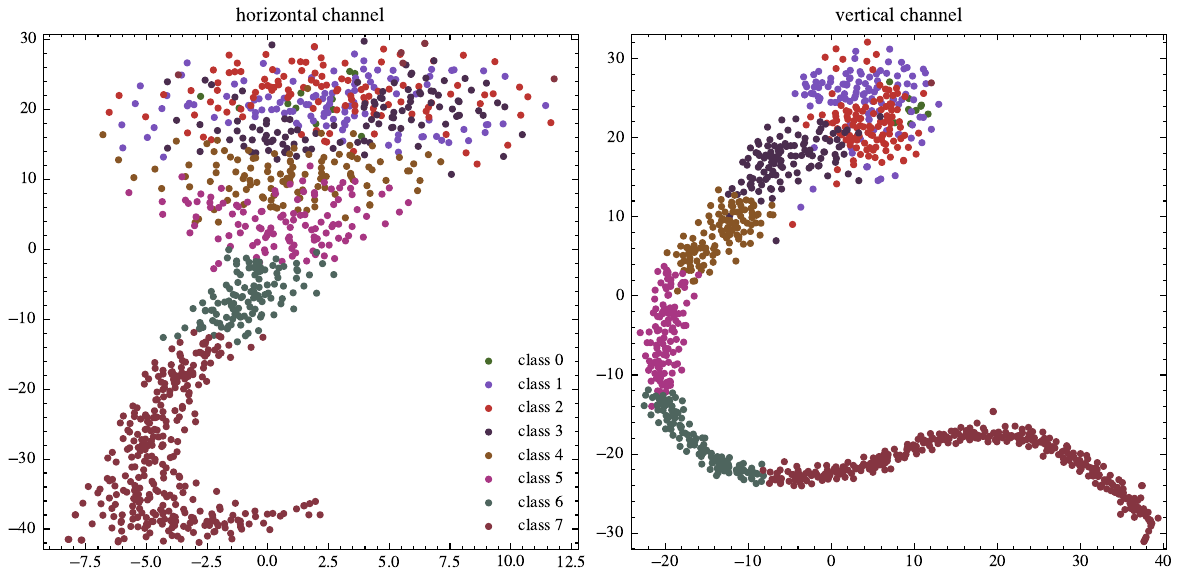} & \includegraphics[height=3.8cm]{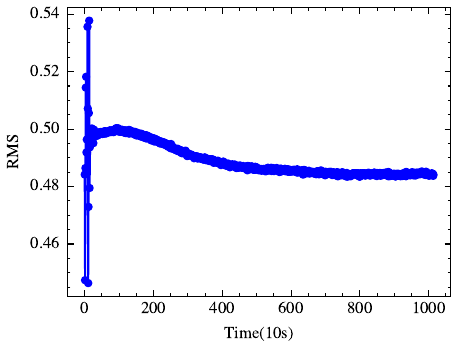} & \includegraphics[height=3.8cm]{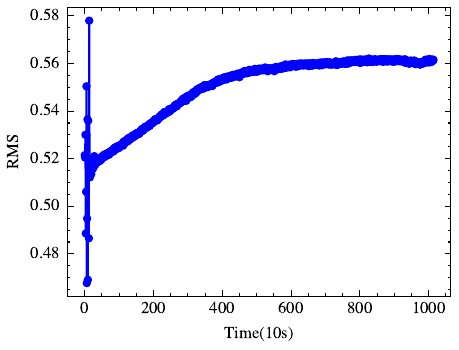} \\
        (a) t-SNE & (b) Horizontal RMS (c) Vertical RMS\\
    \end{tabular}
    \caption{Autoregressive Generation Evaluation Results of CVGAN}
    \label{fig:pred3}
\end{figure*}

The results of these tests, including the root mean square (RMS) and t-SNE analyses, are illustrated in Figures \ref{fig:pred1}-\ref{fig:pred3}.
As corroborated by previous test results, the performance of the WGAN model is observed to be the least effective. This is evident from the continuous increase in the RMS values from t=0 to approximately t=100, where both horizontal and vertical RMS exceed normal ranges. The t-SNE results further reveal a high dispersion in the generated data, with little to no commonality observed.

In contrast, the CVGAN model demonstrates significantly better performance. The RMS fluctuation in the generated data is minimal and within the normal range of data oscillation. This is consistent across the dataset, except for the initial window, where the unconditional generation of our starting model shows suboptimal results. The t-SNE analysis indicates a clear clustering pattern, with each category distinctly identifiable, especially in the vertical direction.

Similarly, the CVAE, another high-performing model, achieves commendable results. The RMS values exhibit no substantial fluctuations, and the t-SNE results, both in horizontal and vertical orientations, are impressive. The CVAE even surpasses CVGAN in terms of accuracy in horizontal direction generation.

In addition to our previous analyses, we employed test models, specifically SCNN and GRU, to perform quantitative assessments of the outputs generated by the autoregressive models. Our evaluation utilized three extensively acknowledged metrics: root mean square error (RMSE), mean absolute error (MAE), and a metric we designate as 'Score'. The formulas for RMSE, MAE, and Score, essential for interpreting our results, are provided as follows:
\begin{equation}\text{RMSE}=\sqrt{\frac{1}{N}\sum_{i=1}^{N}\left(\hat x_i-x_{i}\right)^{2}}\end{equation}
\begin{equation}
\text{MAE}=\frac{1}{N}\sum_{i=1}^{N}|\hat x_i-x_{i}|
\end{equation}
\begin{equation}
\begin{aligned}
E_{i}&=\hat x_{i}-x_{i}\\
A_i&=\begin{cases}
\exp^{-(\frac{E_{i}}{13})} -1
&\text{ if }E_{i}\leq0\\
\exp^{(\frac{E_{i}}{10})}  -1
&\text{ if }E_{i}>0 
\end{cases} \\
Score&=\sum_{i=1}^{N}(A_{i}) 
\end{aligned}
\end{equation}
where $\hat x_i$ represents the real RUL for the $i^{th}$ subject of prediction, while $x_i$ indicates the forecasted RUL for the same predictive subject. Additionally, $N$ symbolizes the total count of samples.

The prediction results of bearing 1-1,1-3 are shown in the Table \ref{tab:pred}. We find that the SCNN model exhibited subpar performance without additional training data, with a MAE exceeding 0.4. On bearing 1-1, the application of the CVGAN method yielded the most favorable results, while the WGAN approach showed negligible improvement. In contrast, for bearing 1-3, the use of extra data generated by any model significantly enhanced performance. However, in the GRU-trained model, only the CVGAN method brought a slight improvement, whereas the CVAE approach was markedly ineffective. The GRU model, when used on bearing 1-1 with any supplemental data, failed to deliver positive outcomes. This indicates that for GRU models, which are already performing well, only high-quality generated data can further enhance predictive performance. Conversely, the SCNN model's inherently poor predictive ability means that any additional training data can improve its performance. Interestingly, even with mediocre quality of extra training data, its predictive accuracy could surpass that of the more robust GRU model.The above experimental results show that our proposed autoregressive generation strategy can significantly enhance the accuracy of RUL prediction, especially in the scenario where the model itself is not strong and the data is very limited.

\begin{table}[]
\centering
\caption{Prediction results using autoregressive generation data}
\label{tab:pred}
\resizebox{\columnwidth}{!}{%
\begin{tabular}{@{}lllllll@{}}
\toprule
               & \multicolumn{3}{c}{SCNN Test on Bearing 1-1}  & \multicolumn{3}{c}{SCNN Test on Bearing 1-3}  \\ \midrule
\textbf{Model} & \textbf{MAE} & \textbf{RMSE} & \textbf{Score} & \textbf{MAE} & \textbf{RMSE} & \textbf{Score} \\
No Extra Data  & 0.450        & 0.503         & 100.591        & 0.422        & 0.444         & 83.256         \\
CVAE           & 0.283        & 0.313         & 63.623         & \textbf{0.106}        & \textbf{0.182}         & \textbf{25.285}         \\
WGAN           & 0.400        & 0.451         & 88.511         & 0.117        & 0.189         & 25.578         \\ 
CVGAN          & \textbf{0.240}        & \textbf{0.278}         & \textbf{61.165}         & 0.136        & 0.218         & 30.202         \\
\midrule
               & \multicolumn{3}{c}{GRU Test on Bearing 1-1}   & \multicolumn{3}{c}{GRU Test on Bearing 1-3}   \\
\textbf{Model} & \textbf{MAE} & \textbf{RMSE} & \textbf{Score} & \textbf{MAE} & \textbf{RMSE} & \textbf{Score} \\
No Extra Data  & \textbf{0.222}        & 0.332         & \textbf{61.852}         & 0.123        & 0.233         & 28.923         \\
CVAE           & 0.263        & \textbf{0.301}         & 66.241         & 0.256        & 0.301         & 53.556         \\
WGAN           & 0.240        & 0.309         & 63.365         & 0.151        & 0.241         & 33.995         \\
CVGAN          & 0.249        & 0.336         & 66.937         & \textbf{0.105}        & \textbf{0.221}         & \textbf{25.316}         \\ \bottomrule
\end{tabular}%
}
\end{table}

\section{Conclusion} \label{section:con}
This paper introduces a CVGAN model capable of directly receiving historical vibration data and category labels, significantly enhancing the generated data's quality. Traditional methods often struggle to generate high-quality data for the entire lifecycle of bearing vibrations. In response, we propose an autoregressive generation method that utilizes historical information to guide the generation process for the current moment. Given the lack of a unified evaluation metric in the academic community for the task of generating 1D vibration signal time series, this study, after thorough analysis and validation, adopts the MMD and FID metrics for this purpose. The experimental results on the PHM 2012 dataset demonstrate that our proposed CVGAN model outperforms many advanced methods in terms of MMD and FID metrics, both in autoregressive and non-autoregressive generation. Training with the full lifecycle data generated by the CVGAN model results in significant performance improvements across multiple prediction models. This outcome highlights the effectiveness of the CVGAN-generated data in enhancing the predictive capabilities of these models.

Future work will focus on enhancing the performance of the initial generator to address the current bottlenecks in generation. Additionally, we plan to explore methods for generating two-dimensional vibration graphs and various more efficient network structures to further narrow the gap between autoregressive and non-autoregressive generation methods.

\section*{Declaration of Competing Interest}
The authors declare that they have no known competing financial interests or personal relationships that could have appeared to influence the work reported in this paper.

\section*{Acknowledgments}
The research was partially supported by the Key Project of Natural Science Foundation of China (No. 61933013), the Science and Technology Innovation Strategy Project of Guangdong Provincial(No. 2023S002028) and the special projects in key fields of ordinary universities in Guangdong Province(No.2023ZDZX3015). We also appreciate the technical support of the Guangdong Provincial Key Laboratory of Petrochemical Equipment and Fault Diagnosis.



\bibliographystyle{elsarticle-num} 
\bibliography{paper}

\begin{thebibliography}{10}
\expandafter\ifx\csname url\endcsname\relax
  \def\url#1{\texttt{#1}}\fi
\expandafter\ifx\csname urlprefix\endcsname\relax\def\urlprefix{URL }\fi
\expandafter\ifx\csname href\endcsname\relax
  \def\href#1#2{#2} \def\path#1{#1}\fi

\bibitem{physical1}
Y.~Song, S.~Xu, X.~Lu, A sliding sequence importance resample filtering method for rolling bearings remaining useful life prediction based on two wiener-process models, Measurement Science and Technology 35~(1) (2023) 015019.
\newblock \href {https://doi.org/10.1088/1361-6501/acffe3} {\path{doi:10.1088/1361-6501/acffe3}}.

\bibitem{physical2}
S.~M.~M. Hassani.N, X.~Jin, J.~Ni, Physics-based gaussian process for the health monitoring for a rolling bearing, Acta Astronautica 154 (2019) 133--139.
\newblock \href {https://doi.org/10.1016/j.actaastro.2018.10.029} {\path{doi:10.1016/j.actaastro.2018.10.029}}.

\bibitem{hybrid1}
B.~Wang, Y.~Lei, N.~Li, J.~Lin, An improved fusion prognostics method for remaining useful life prediction of bearings, in: 2017 IEEE International Conference on Prognostics and Health Management (ICPHM), 2017, pp. 18--24.
\newblock \href {https://doi.org/10.1109/ICPHM.2017.7998300} {\path{doi:10.1109/ICPHM.2017.7998300}}.

\bibitem{hybrid2}
D.~Liu, W.~Xie, S.~Lu, Y.~Peng, Battery prognostics with uncertainty fusion for aerospace applications, in: 2015 Annual Reliability and Maintainability Symposium (RAMS), 2015, pp. 1--6.
\newblock \href {https://doi.org/10.1109/RAMS.2015.7105073} {\path{doi:10.1109/RAMS.2015.7105073}}.

\bibitem{7}
H.~Miao, B.~Li, C.~Sun, J.~Liu, Joint learning of degradation assessment and rul prediction for aeroengines via dual-task deep lstm networks, IEEE Transactions on Industrial Informatics 15~(9) (2019) 5023--5032.
\newblock \href {https://doi.org/10.1109/TII.2019.2900295} {\path{doi:10.1109/TII.2019.2900295}}.

\bibitem{8}
Z.~Chen, K.~Gryllias, W.~Li, Mechanical fault diagnosis using convolutional neural networks and extreme learning machine, Mechanical Systems and Signal Processing 133 (2019) 106272.
\newblock \href {https://doi.org/10.1016/j.ymssp.2019.106272} {\path{doi:10.1016/j.ymssp.2019.106272}}.

\bibitem{re5}
W.~Mao, J.~He, J.~Tang, Y.~Li, Predicting remaining useful life of rolling bearings based on deep feature representation and long short-term memory neural network, Advances in Mechanical Engineering 10~(12) (2018) 1687814018817184.
\newblock \href {https://doi.org/10.1177/1687814018817184} {\path{doi:10.1177/1687814018817184}}.

\bibitem{1-1}
X.~Li, K.~Zhang, W.~Li, Y.~Feng, R.~Liu, A two-stage transfer regression convolutional neural network for bearing remaining useful life prediction, Machines 10~(5) (2022) 369.
\newblock \href {https://doi.org/10.3390/machines10050369} {\path{doi:10.3390/machines10050369}}.

\bibitem{1-2}
J.~Yang, Y.~Peng, J.~Xie, P.~Wang, Remaining useful life prediction method for bearings based on lstm with uncertainty quantification, Sensors 22~(12) (2022) 4549.
\newblock \href {https://doi.org/10.3390/s22124549} {\path{doi:10.3390/s22124549}}.

\bibitem{1-3}
Q.~Chen, X.~Ma, B.~Yan, W.~Yanyan, G.~Huang, Remaining useful life prediction of bearings with two-stage lstm, 2022 5th International Symposium on Autonomous Systems (ISAS) (2022) 1--6\href {https://doi.org/10.1109/ISAS55863.2022.9757261} {\path{doi:10.1109/ISAS55863.2022.9757261}}.

\bibitem{1-4}
R.~Guo, Y.~Wang, H.~Zhang, G.~Zhanga, Remaining useful life prediction for rolling bearings using emd-risi-lstm, IEEE Transactions on Instrumentation and Measurement (2021) 1--1\href {https://doi.org/10.1109/TIM.2021.3051717} {\path{doi:10.1109/TIM.2021.3051717}}.

\bibitem{small1}
G.~Jin, D.~Matthews, Y.~Fan, Q.~Liu, Physics of failure-based degradation modeling and lifetime prediction of the momentum wheel in a dynamic covariate environment, Engineering Failure Analysis 28 (2013) 222--240.
\newblock \href {https://doi.org/10.1016/j.engfailanal.2012.10.027} {\path{doi:10.1016/j.engfailanal.2012.10.027}}.

\bibitem{small2}
C.~Lu, J.~Chen, R.~Hong, Y.~Feng, Y.~Li, Degradation trend estimation of slewing bearing based on lssvm model, Mechanical Systems and Signal Processing 76--77 (2016) 353--366.
\newblock \href {https://doi.org/10.1016/j.ymssp.2016.02.031} {\path{doi:10.1016/j.ymssp.2016.02.031}}.

\bibitem{small3}
Z.~Pan, Z.~Meng, Z.~Chen, W.~Gao, Y.~Shi, A two-stage method based on extreme learning machine for predicting the remaining useful life of rolling-element bearings, Mechanical Systems and Signal Processing 144 (2020) 106899.
\newblock \href {https://doi.org/10.1016/j.ymssp.2020.106899} {\path{doi:10.1016/j.ymssp.2020.106899}}.

\bibitem{small4}
W.~Chen, W.~Chen, H.~Liu, Y.~Wang, C.~Bi, Y.~Gu, A rul prediction method of small sample equipment based on dcnn-bilstm and domain adaptation, Mathematics 10~(7) (2022) 1022.
\newblock \href {https://doi.org/10.3390/math10071022} {\path{doi:10.3390/math10071022}}.

\bibitem{gan}
I.~Goodfellow, J.~{Pouget-Abadie}, M.~Mirza, B.~Xu, D.~{Warde-Farley}, S.~Ozair, A.~Courville, Y.~Bengio, Generative adversarial nets, in: Advances in Neural Information Processing Systems, Vol.~27, Curran Associates, Inc., 2014.

\bibitem{wgan}
M.~Arjovsky, S.~Chintala, L.~Bottou, Wasserstein generative adversarial networks, in: Proceedings of the 34th International Conference on Machine Learning, PMLR, 2017, pp. 214--223.

\bibitem{wgan-gp}
I.~Gulrajani, F.~Ahmed, M.~Arjovsky, V.~Dumoulin, A.~C. Courville, Improved training of wasserstein gans, in: Advances in Neural Information Processing Systems, Vol.~30, Curran Associates, Inc., 2017.

\bibitem{bigan}
J.~Donahue, P.~Kr{\"a}henb{\"u}hl, T.~Darrell, Adversarial feature learning (2017).
\newblock \href {http://arxiv.org/abs/1605.09782} {\path{arXiv:1605.09782}}, \href {https://doi.org/10.48550/arXiv.1605.09782} {\path{doi:10.48550/arXiv.1605.09782}}.

\bibitem{infogan}
X.~Chen, Y.~Duan, R.~Houthooft, J.~Schulman, I.~Sutskever, P.~Abbeel, Infogan: Interpretable representation learning by information maximizing generative adversarial nets, in: Advances in Neural Information Processing Systems, Vol.~29, Curran Associates, Inc., 2016.

\bibitem{2-1}
J.~Man, M.~Zheng, Y.~Liu, Y.~Shen, Q.~Li, Bearing remaining useful life prediction based on adcnn and cwgan under few samples, Shock and Vibration 2022 (2022) 1--17.
\newblock \href {https://doi.org/10.1155/2022/1709071} {\path{doi:10.1155/2022/1709071}}.

\bibitem{2-2}
S.~Wang, J.~Yu, K.~Zhao, Z.~Guo, Bearing remaining life prediction method based on wavelet transform denoising and feature preservation cyclegan under unbalanced samples, 2022 International Conference on Sensing, Measurement \& Data Analytics in the era of Artificial Intelligence (ICSMD) (2022) 1--6\href {https://doi.org/10.1109/ICSMD57530.2022.10058294} {\path{doi:10.1109/ICSMD57530.2022.10058294}}.

\bibitem{2-3}
S.~Zhang, T.~Li, X.~Si, C.~Hu, H.~Zhang, Y.~Ma, A new missing data generation method based on an improved dcgan with application to rul prediction, 2021 CAA Symposium on Fault Detection, Supervision, and Safety for Technical Processes (SAFEPROCESS) (2021) 1--6\href {https://doi.org/10.1109/SAFEPROCESS52771.2021.9693658} {\path{doi:10.1109/SAFEPROCESS52771.2021.9693658}}.

\bibitem{vae}
D.~P. Kingma, M.~Welling, Auto-encoding variational bayes (2022).
\newblock \href {http://arxiv.org/abs/1312.6114} {\path{arXiv:1312.6114}}, \href {https://doi.org/10.48550/arXiv.1312.6114} {\path{doi:10.48550/arXiv.1312.6114}}.

\bibitem{vae1}
J.~Peng, D.~Liu, S.~Xu, H.~Li, Generating diverse structure for image inpainting with hierarchical vq-vae, in: Proceedings of the IEEE/CVF Conference on Computer Vision and Pattern Recognition, 2021, pp. 10775--10784.

\bibitem{vqvae}
A.~{van den Oord}, O.~Vinyals, k.~{kavukcuoglu}, Neural discrete representation learning, in: Advances in Neural Information Processing Systems, Vol.~30, Curran Associates, Inc., 2017.

\bibitem{vae2}
X.~Li, J.~She, Collaborative variational autoencoder for recommender systems, in: Proceedings of the 23rd ACM SIGKDD International Conference on Knowledge Discovery and Data Mining, KDD '17, Association for Computing Machinery, New York, NY, USA, 2017, pp. 305--314.
\newblock \href {https://doi.org/10.1145/3097983.3098077} {\path{doi:10.1145/3097983.3098077}}.

\bibitem{vae3}
Y.~Shi, S.~N, B.~Paige, P.~Torr, Variational mixture-of-experts autoencoders for multi-modal deep generative models, in: Advances in Neural Information Processing Systems, Vol.~32, Curran Associates, Inc., 2019.

\bibitem{cvae-gan}
J.~Bao, D.~Chen, F.~Wen, H.~Li, G.~Hua, Cvae-gan: Fine-grained image generation through asymmetric training, in: 2017 IEEE International Conference on Computer Vision (ICCV), IEEE, Venice, 2017, pp. 2764--2773.
\newblock \href {https://doi.org/10.1109/ICCV.2017.299} {\path{doi:10.1109/ICCV.2017.299}}.

\bibitem{ar1}
K.~Chen, G.~Chen, D.~Xu, L.~Zhang, Y.~Huang, A.~Knoll, Nast: Non-autoregressive spatial-temporal transformer for time series forecasting, https://arxiv.org/abs/2102.05624v2 (2021).

\bibitem{ar2}
T.~Brown, B.~Mann, N.~Ryder, M.~Subbiah, J.~D. Kaplan, P.~Dhariwal, A.~Neelakantan, P.~Shyam, G.~Sastry, A.~Askell, S.~Agarwal, A.~{Herbert-Voss}, G.~Krueger, T.~Henighan, R.~Child, A.~Ramesh, D.~Ziegler, J.~Wu, C.~Winter, C.~Hesse, M.~Chen, E.~Sigler, M.~Litwin, S.~Gray, B.~Chess, J.~Clark, C.~Berner, S.~McCandlish, A.~Radford, I.~Sutskever, D.~Amodei, Language models are few-shot learners, Advances in Neural Information Processing Systems 33 (2020) 1877--1901.

\bibitem{ar2-2}
J.~Ma, F.~Xu, K.~Huang, R.~Huang, Gnar-garch model and its application in feature extraction for rolling bearing fault diagnosis, Mechanical Systems and Signal Processing 93 (2017) 175--203.
\newblock \href {https://doi.org/10.1016/j.ymssp.2017.01.043} {\path{doi:10.1016/j.ymssp.2017.01.043}}.

\bibitem{ar4}
V.~M. Nistane, Wavelet-based features for prognosis of degradation in rolling element bearing with non-linear autoregressive neural network, Australian Journal of Mechanical Engineering 19~(4) (2021) 423--437.
\newblock \href {https://doi.org/10.1080/14484846.2019.1630949} {\path{doi:10.1080/14484846.2019.1630949}}.

\bibitem{33}
R.~Wang, R.~Shi, X.~Hu, C.~Shen, Remaining useful life prediction of rolling bearings based on multiscale convolutional neural network with integrated dilated convolution blocks, Shock and Vibration 2021 (2021) e6616861.
\newblock \href {https://doi.org/10.1155/2021/6616861} {\path{doi:10.1155/2021/6616861}}.

\bibitem{34}
Y.~Cao, M.~Jia, P.~Ding, Y.~Ding, Transfer learning for remaining useful life prediction of multi-conditions bearings based on bidirectional-gru network, Measurement 178 (2021) 109287.
\newblock \href {https://doi.org/10.1016/j.measurement.2021.109287} {\path{doi:10.1016/j.measurement.2021.109287}}.

\bibitem{dataset}
P.~Nectoux, R.~Gouriveau, K.~Medjaher, E.~Ramasso, B.~{Chebel-Morello}, N.~Zerhouni, C.~Varnier, Pronostia : An experimental platform for bearings accelerated degradation tests., in: IEEE International Conference on Prognostics and Health Management, PHM'12., Vol. sur CD ROM, IEEE Catalog Number : CPF12PHM-CDR, Denver, Colorado, United States, 2012, pp. 1--8.

\bibitem{fpt}
N.~Li, Y.~Lei, J.~Lin, S.~X. Ding, An improved exponential model for predicting remaining useful life of rolling element bearings, IEEE Transactions on Industrial Electronics 62~(12) (2015) 7762--7773.
\newblock \href {https://doi.org/10.1109/TIE.2015.2455055} {\path{doi:10.1109/TIE.2015.2455055}}.

\bibitem{our2}
J.~Wang, Q.~Zhang, G.~Zhu, G.~Sun, Utilizing vq-vae for end-to-end health indicator generation in predicting rolling bearing rul, https://arxiv.org/abs/2311.10525v1 (2023).

\bibitem{aae-gan}
A.~Makhzani, J.~Shlens, N.~Jaitly, I.~Goodfellow, B.~Frey, Adversarial autoencoders (2016).
\newblock \href {http://arxiv.org/abs/1511.05644} {\path{arXiv:1511.05644}}, \href {https://doi.org/10.48550/arXiv.1511.05644} {\path{doi:10.48550/arXiv.1511.05644}}.

\end{thebibliography}





% \end{thebibliography}
\end{document}